%% file: main.tex
\documentclass{article}

\usepackage{arxiv}

\usepackage[utf8]{inputenc} 
\usepackage[T1]{fontenc}    
\usepackage{hyperref}       
\usepackage{url}            
\usepackage{booktabs}       
\usepackage{amsfonts}       
\usepackage{nicefrac}       
\usepackage{microtype}      
\usepackage{lipsum}		
\usepackage{graphicx}
\usepackage{natbib}
\usepackage{doi}

\usepackage{booktabs}
\usepackage[table]{xcolor}
\usepackage{graphicx}
\usepackage{multirow}
\usepackage{colortbl}
\usepackage{makecell}
\definecolor{LightGreen}{RGB}{200,255,200}
\definecolor{LightRed}{RGB}{255,200,200}
\usepackage{algorithm2e}
\usepackage{enumitem}
\usepackage{comment}

\usepackage{amsthm}   
\usepackage{amsmath,amssymb}
\usepackage{mathtools}

\usepackage{enumitem}

\theoremstyle{plain}
\newtheorem{theorem}{Theorem}[section]
\newtheorem{proposition}[theorem]{Proposition}

\theoremstyle{definition}

\theoremstyle{remark}

\usepackage{amssymb}

\usepackage{hyperref}

\usepackage{booktabs}
\usepackage[table]{xcolor}
\usepackage{graphicx}
\usepackage{multirow}
\usepackage{colortbl}
\usepackage{makecell}
\definecolor{LightGreen}{RGB}{200,255,200}
\definecolor{LightRed}{RGB}{255,200,200}
\usepackage{algorithm2e}
\usepackage{amsmath}
\usepackage{amsthm}

\usepackage{pifont}

\usepackage{etoolbox}

\preto\appendix{%
}

\preto\appendix{%
}

\title{\textsc{MissionHD}: Hyperdimensional Refinement of Distribution-Deficient Reasoning Graphs for Video Anomaly Detection}

\date{} 					

\author{
    Sanggeon Yun\\
    Department of Computer Science\\
    University of California, Irvine\\
    CA, USA\\
    \texttt{sanggeoy@uci.edu} \\
\And
    Raheeb Hassan\\
    Department of Computer Science\\
    University of California, Irvine\\
    CA, USA\\
    \texttt{raheebh@uci.edu} \\
\And
    Ryozo Masukawa\\
    Department of Computer Science\\
    University of California, Irvine\\
    CA, USA\\
    \texttt{rmasukaw@uci.edu} \\
\And
    Nathaniel D. Bastian\\
    Department of Electrical Engineering \& Computer Science\\
    United States Military Academy, West Point\\
    NY, USA\\
    \texttt{nathaniel.bastian@westpoint.edu} \\
\And
    Mohsen Imani\\
    Department of Computer Science\\
    University of California, Irvine\\
    CA, USA\\
    \texttt{m.imani@uci.edu} \\
}

\date{}

\renewcommand{\shorttitle}{\textsc{MissionHD}: Hyperdimensional Refinement of Distribution-Deficient Graphs for VAD}
\newcommand{\MissionHD}{\textsc{MissionHD}}

\hypersetup{
pdftitle={MissionHD: Hyperdimensional Refinement of Distribution-Deficient Reasoning Graphs for Video Anomaly Detection},
pdfauthor={Sanggeon Yun},
}

\begin{document}
\maketitle

\input{Sections/0_Abstract}


\section{Introduction}
\input{Sections/1_Introduction}

\section{Preliminaries \& Related Works}
\input{Sections/2_Preliminaries}

\section{Methodology}
\input{Sections/3_Methodology}

\section{Experiments}
\label{sec:experiments}
\input{Sections/4_Experiments}

\section{Conclusions}
\input{Sections/6_Conclusions}

\section*{Acknowledgements}
This work was supported in part by the DARPA Young Faculty Award, the National Science Foundation (NSF) under Grants \#2127780, \#2319198, \#2321840, \#2312517, \#2235472, and \#2431561, the Semiconductor Research Corporation (SRC), the Office of Naval Research through the Young Investigator Program Award, and Grants \#N00014-21-1-2225 and \#N00014-22-1-2067, Army Research Office Grant \#W911NF2410360. Additionally, support was provided by the Air Force Office of Scientific Research under Award \#FA9550-22-1-0253, along with generous gifts from Xilinx and Cisco.

\newpage
\bibliographystyle{plain}
\bibliography{ref.bib}

\newpage
\appendix

\section*{Appendix}

\section{Theoretical Analysis}\label{appx:thereticalanalysis}

\paragraph{Assumptions.}\label{assump:assumptions}
Our analysis relies on standard HDC assumptions:
\begin{enumerate}[label=(A\arabic*), leftmargin=2em]
    \item \textbf{Near-orthogonality:} For random vectors $u, v \in \mathbb{R}^D$ with i.i.d. zero-mean, unit-variance components, $\mathbb{E}[u^\top v]=0$. By concentration inequalities (e.g., Hoeffding's), for any $\tau>0$, $\Pr(|\frac{1}{D}u^\top v| > \tau) \le 2\exp(-c D \tau^2)$ for some constant $c>0$.
    \item \textbf{Similarity Preservation:} Binding approximately preserves cosine similarity, i.e., $\delta(q\otimes a, q\otimes b) \approx \delta(a,b)$.
    \item \textbf{Johnson–Lindenstrauss Preservation:} The projection $\phi$ approximately preserves pairwise distances and angles.
\end{enumerate}

\subsection{HDC-Constrained GNN Equivalence}

\paragraph{\autoref{thm:hdcgnneq}}
\begin{proof}
By definition of element-wise multiplication, $[L^{(i)}\otimes z]_d = L^{(i)}_d z_d = [\mathrm{diag}(L^{(i)}) z]_d$ for each coordinate $d$.
\end{proof}

\paragraph{\autoref{thm:pathsumgnn}}
\begin{proof}
A linearized GNN update is $h_k^{(i)} = W^{(i)} \sum_{t \in N^-(v_k^{(i)})} h_t^{(i-1)}$ where $N^-(v_k^{(i)})$ denotes the set of in-neighbors of $v_k^{(i)}$. With HDC constraints, this becomes $h_k^{(i)} = D_{L^{(i)}} \bigoplus_{t} (H_t^{(i-1)} \otimes h_t^{(i-1)})$. Unrolling this recurrence from layer $i$ to 1 yields a sum-of-products expression. As all operators are diagonal matrices (bindings) or sums (bundlings), the GNN output becomes a grand bundle of bindings along each path from source to sink, which is precisely what the DP algorithm for $H_G$ computes.
\end{proof}

\subsection{Softmax selection at threshold $T$}

Let $s_{i,k,t}$ be the softmax-normalized edge score from \autoref{eq:edgescore} and fix a threshold $T\in(0,1)$ as in \autoref{alg:decoding}.
Define the margin $\gamma_{i,k,t} := S_{i,k,t} - \max_{t'\neq t} S_{i,k,t'}$.
If $\mu_{i,k,t} - \max_{t'\neq t}\mu_{i,k,t'} \ge \gamma>0$, then, for $\Delta_T := \gamma - \log\!\frac{T}{1-T}$,
\begin{align*}
\Pr\!\big(s_{i,k,t} \le T\big)
\;\le\;
2\exp\!\left(-\,\frac{D\,\Delta_T^2}{K_{i,k}}\right),
\qquad\\
K_{i,k}:= 2c\Big(\,|P_{\neg(k\to t)}| + \max_{t'\neq t}|P_{\neg(k\to t')}|\,\Big),
\end{align*}
i.e., the chance that a truly contributing edge falls below threshold decays exponentially in $D$ and the softmax margin.

\subsection{Decoding guarantees}

\paragraph{\autoref{prop:decode-accuracy-strong}}
\begin{proof}
Write $S(\mathcal{T}):=\mathrm{span}\!\big(\{H_{\text{path}}(p):\,p\in \mathcal{P}_{\text{keep}}(\mathcal{T})\}\cup \{H_{\text{path}}(q):\,q\in Q\}\big)$, and let $\Pi_{S(\mathcal{T})}$ be the orthogonal projector onto $S(\mathcal{T})$.
Decompose the target unit vector as
\begin{align*}
\widehat{H}'_G \;=\; u \;+\; r, 
\quad u:=\Pi_{S(\mathcal{T})}\widehat{H}'_G \in S(\mathcal{T}), \\
\quad r\perp S(\mathcal{T}), 
\quad \|u\|_2^2+\|r\|_2^2=1.
\end{align*}
The refined re-encoding $H_G^{\text{refined}}$ lies in $S(\mathcal{T})$ (it re-bundles only kept and new paths), so
\(
\delta(H_G^{\text{refined}},H'_G)
= \langle \widehat{H}_G^{\text{refined}}, \widehat{H}'_G\rangle
= \langle \widehat{H}_G^{\text{refined}}, u\rangle
\).
Maximizing over all unit vectors in $S(\mathcal{T})$ gives 
\(
\delta(H_G^{\text{refined}},H'_G) \le \|u\|_2
\),
and the Pythagorean identity yields
\[
1-\big(\delta(H_G^{\text{refined}},H'_G)\big)^2
\;\le\;
1-\|u\|_2^2
\;=\;
\|r\|_2^2. \tag{$\star$}
\]
Next split $r$ into two orthogonal parts: (i) residual from \emph{discarded original paths}, and (ii) residual from \emph{edit information} not captured by $Q$.
For a discarded path $p\in\mathcal{P}_{\text{miss}}(\mathcal{T})$, at least one of its edges has softmax score $\le \mathcal{T}$; by \autoref{alg:decoding} and the multiplicative nature of HDC composition along a path, its contribution to $\langle H_{\text{path}}(p), \widehat{H}'_G\rangle$ is geometrically damped along the chain.
Empirically, this gives a per-path bound
\(
|\langle H_{\text{path}}(p), \widehat{H}'_G\rangle|
= O\big((\tau_{\max}(\mathcal{T}))^{L(p)}\big)
\),
hence the total energy of all discarded-path components satisfies
\[
\big\|r_{\text{miss}}\big\|_2^2
\;=\;
O\!\Big(\sum_{p\in \mathcal{P}_{\text{miss}}(\mathcal{T})} (\tau_{\max}(\mathcal{T}))^{2L(p)}\Big).
\]
For the edit vector, decompose $\widehat{H}_\mathcal{E}$ into its projection onto $\mathrm{span}(\{H_{\text{path}}(q):\,q\in Q\})$ and its orthogonal complement; the latter is exactly the \emph{new-path approximation} residual
\(
\big\|\,\widehat{H}_\mathcal{E} - \Pi_{\mathrm{span}(Q)}\widehat{H}_\mathcal{E}\,\big\|_2
\),
which adds additively to $\|r\|_2$.
Finally, finite-$D$ crosstalk among the $(|\mathcal{P}_{\text{keep}}(\mathcal{T})|+|Q|)$ bundled codes contributes an additional variance term that vanishes as $D\uparrow$ by (A1).
Standard concentration for sums of near-orthogonal random hypervectors in $\mathbb{R}^D$ (Hoeffding-type bounds over coordinates) gives a mean-square leakage of order $O\big((|\mathcal{P}_{\text{keep}}(\mathcal{T})|+|Q|)/D\big)$ into $r$.
Combining these three pieces with $(\star)$ proves the stated Big-$O$ bound.
\qedhere
\end{proof}


\section{Detailed Algorithm}

Pseudocode for the proposed hyperdimensional encoding and decoding methods is provided in \autoref{alg:encoding}, \autoref{alg:denseencoding}, and \autoref{alg:decoding}.

\begin{algorithm}[h]
\caption{Hyperdimensional reasoning graph encoding. Encodes graph paths efficiently into a global hypervector $H_G$ using DP. The $\mathrm{Norm}(\cdot)$ function denotes element-wise normalization to unit length, ensuring numerical stability.}
\label{alg:encoding}
\DontPrintSemicolon
\SetKwInOut{Input}{Input}
\SetKwInOut{Output}{Output}

\Input{
  Set of vertices: $\mathcal{V}$,\;
  Set of directed edges $(v\!\to\!u)$: $\mathcal{E}$,\;
  Node hypervectors: $\{H^{(d)}_k\}$,\;
  Layer hypervectors: $\{L^{(d)}\}$,\;
  Graph edit hypervectors: $H_\mathcal{E}$,\;
  Depath sets: $\{\mathcal{V}_d\}$
}
\Output{
  Encoded graph hypervector $H_G$
}

$M^{(0)}_{\rm k}\gets \mathbf{1}$\;
\For{$d=1, \dots, \max(\{d\})$}{
  \For{$v^{d}_k\in \mathcal{V}_d$}{
    $M^{(d)}_{k}\gets \bigoplus_{v_t^{(d-1)}\rightarrow v_k^{(d)}}\mathrm{Norm}\bigl(L^{(d-1)}\otimes F_t^{(d-1)}\otimes H_t^{(d-1)}\bigr)$\;
    $M^{(d)}_{k}\gets M^{(d)}_{k}/|\{v_t^{(d-1)}\colon v_t^{(d-1)}\rightarrow v_k^{(d)}\}|$\;
  }
}
$H_G=\bigoplus_{v_k^{(\ell)}}L^{(\ell)}\otimes M_k^{(\ell)}\otimes H_k^{(\ell)}$\;
\Return{$H_G$}
\end{algorithm}

\begin{algorithm}[h]
\caption{Hyperdimensional Encoding Forward \& Reverse. Computes forward and backward messages under full connectivity assumptions for evaluating potential edge contributions.}
\label{alg:denseencoding}
\DontPrintSemicolon
\SetKwInOut{Input}{Input}
\SetKwInOut{Output}{Output}

\Input{
  $\mathcal{V},\mathcal{E},\{H^{(d)}_k\}, \{L^{(d)}\}, H_\mathcal{E}, \{\mathcal{V}_d\}$,\;
}
\Output{
  Forward messages $F$,\;
  Reverse messages $B$
}

\tcp{Forward DP}
$F^{(0)}_{\rm k}\gets \mathbf{1}$\;
\For{$d=1, \dots, \max(\{d\})$}{
  \For{$v^{d}_k\in \mathcal{V}_d$}{
    $F^{(d)}_{k}\gets \bigoplus_{v^{(d-1)}_t\in \mathcal{V}_{d-1}}\mathrm{Norm}\bigl(L^{(d-1)}\otimes F_t^{(d-1)}\otimes H_t^{(d-1)}\bigr)$\;
    $F^{(d)}_{k}\gets F^{(d)}_{k}/|\mathcal{V}_{d-1}|$\;
  }
}

\tcp{Reverse DP}
$B^{(\max(\{d\})+1)}_{\rm k}\gets \mathbf{1}$\;
\For{$d=\max(\{d\}), \dots, 1$}{
  \For{$v^{d}_k\in \mathcal{V}_d$}{
    $B^{(d)}_{k}\gets \bigoplus_{v^{(d+1)}_t\in \mathcal{V}_{d+1}}\mathrm{Norm}\bigl(L^{(d+1)}\otimes B_t^{(d+1)}\otimes H_t^{(d+1)}\bigr)$\;
    $B^{(d)}_{k}\gets B^{(d)}_{k}/|\mathcal{V}_{d+1}|$\;
  }
}

\Return{$F, B$}
\end{algorithm}
\vspace{-2mm}

\begin{algorithm}[h]
\caption{Graph Decoding \& Refinement. Refines the graph structure by estimating edge contribution scores based on compositional similarity between hypothetical edge hypervectors ($FB_{d,k,t}$) and the trained hypervector $H_G'$. Edges surpassing a threshold $\mathcal{T}$ constitute the refined graph structure.}
\label{alg:decoding}
\DontPrintSemicolon
\SetKwInOut{Input}{Input}
\SetKwInOut{Output}{Output}

\Input{
  $\mathcal{V},\mathcal{E},\{H^{(d)}_k\}, \{L^{(d)}\}, H_\mathcal{E}, \{\mathcal{V}_d\}$,\;
  $H_G$ from Alg.~\ref{alg:encoding},\;
  $F, B$ from Alg.~\ref{alg:denseencoding},\;
  threshold $\mathcal{T}$
}
\Output{Refined edge set $\mathcal{E}_{\rm refined}$}

\BlankLine
\textbf{Initialize:} 
\quad Candidates $\leftarrow []$, 
Scores $\leftarrow []$\;
$H_G'\gets H_\mathcal{E}\oplus H_G$\;
\For{$d=1,\dots,\max(\{d\})-1$}{
  \For{$v^{(d)}_k\in \mathcal{V}_d,\;u^{(d+1)}_t\in \mathcal{V}_{d+1}$}{
    $FB_{d, k, t}\gets L^{(d)} \otimes F_k^{(d)} \otimes H_k^{(d)} \otimes L^{(d+1)} \otimes F_t^{(d+1)} \otimes H_t^{(d+1)}$\;
  }
  $s^{d}_{k, t}\gets \mathrm{softmax}_{k, t}\left(\delta(FB(d,k,t), H_G')\right)$\;
  Append $(v^{(d)}_k,u^{(d+1)}_t)$ to Candidates\;
  Append $s^{d}_{k, t}$ to Scores\;
}

$\mathcal{E}_{\rm refined}\gets\{\, (v,u)\in\text{Candidates} : s_{vu}>\mathcal{T}\}\,$

\Return{$\mathcal{E}_{\rm refined}$}
\end{algorithm}

\section{Empirical Analysis of Task-Driven Graph Refinement}
\label{sec:graph_refinement}

\begin{figure*}[h]
    \centering
    \includegraphics[width=1.0\textwidth]{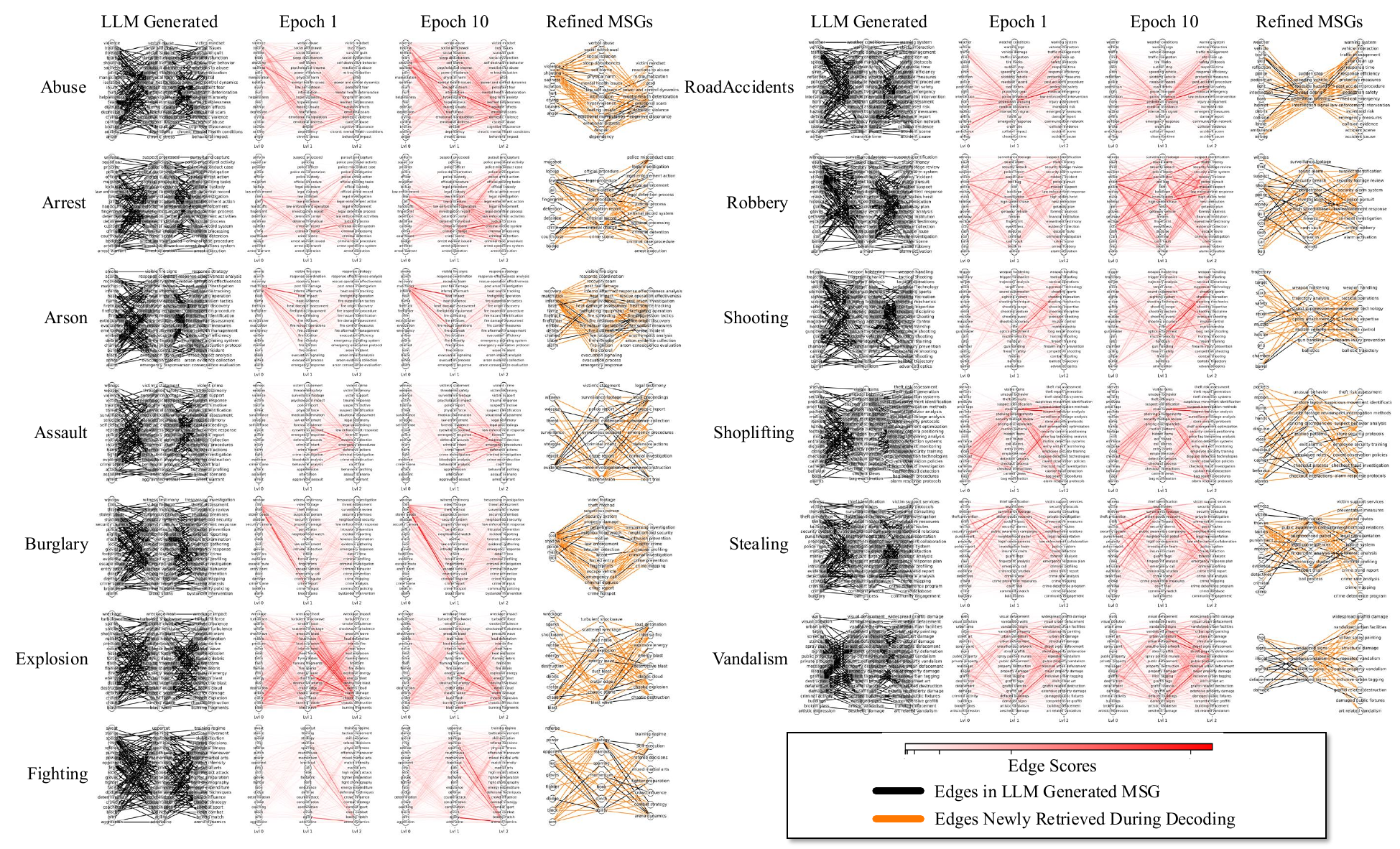}
    \caption{Refinement of LLM-generated knowledge graphs across UCF-Crime classes. Each panel visualizes the evolution of reasoning graph structure from the initial LLM-generated graph (left) through intermediate refinement steps at Epoch 1 and Epoch 10, to the final refined graph (right). This is the first refinement pass, using a threshold value of 0.1 for edge contribution scores. Interestingly, even by Epoch 1, the model begins to recover key structures that appear in the final refined graph, albeit with lower confidence. As training progresses, contribution scores become more polarized, resulting in stronger, more discriminative edge selections. Colored edges represent both original and newly retrieved connections based on our proposed hyperdimensional decoding.
}\label{fig:refinement_figure}
\end{figure*}

A core contribution of our work is the ability of \MissionHD\ to refine the structure of an initial LLM-generated knowledge graph, tailoring it to the specific distribution of data characteristics of a downstream task. This refinement process is critical because LLMs often generate graphs containing generalized concepts without knowing the actual downstream task's data distribution, resulting in abstract concepts or relationships that are difficult to ground in actual downstream data, particularly from sources like surveillance cameras. In this section, we analyze the qualitative changes in the graph structure to demonstrate that our refinement process produces graphs that are more semantically aligned with visual evidence.

\autoref{fig:refinement_figure} visually depicts this evolution for several anomaly classes. The initial LLM-generated graphs are dense and contain many nodes that represent internal states, emotions, or complex abstractions that are far from solving the downstream task. As the model trains, the hyperdimensional decoding mechanism evaluates and prunes irrelevant nodes and edges while strengthening connections that are more predictive for the task. This results in a sparser, more interpretable, and more effective reasoning structure, creating mission-specific graphs that are more relevant to the downstream dataset.

\begin{table}[t!]
\centering
\caption{Qualitative analysis of pruned versus retained nodes for classes highly relevant to CCTV footage. The refinement process systematically removes general or abstract concepts in favor of concrete, observable objects and actions.}
\label{tab:node_analysis}
\resizebox{0.8\columnwidth}{!}{%
\begin{tabular}{lll}
\toprule
\textbf{Class} & \textbf{Category} & \textbf{Nodes} \\
\midrule
\multirow{3}{*}{\textbf{Robbery}} & Pruned (General / Abstract) & \makecell[l]{\texttt{theft, danger, illegal, crime, fear}} \\
\cmidrule(lr){2-3}
& Retained (Specific Objects / Actions) & \makecell[l]{\texttt{weapon, mask, getaway car, forced entry,}\\ \texttt{cash register, demanding money}} \\
\midrule
\multirow{3}{*}{\textbf{Vandalism}} & Pruned (General / Abstract) & \makecell[l]{\texttt{damage, destruction, malicious, illegal act}} \\
\cmidrule(lr){2-3}
& Retained (Specific Objects / Actions) & \makecell[l]{\texttt{graffiti, spray paint, breaking windows,}\\ \texttt{smashing, keying car}} \\
\bottomrule
\end{tabular}%
}
\end{table}

\begin{table}[t!]
\centering
\caption{Comparison of high-scoring and low-scoring edge contributions for the \textit{Robbery} class. High-scoring edges connect specific visual cues to their logical implications, while low-scoring edges represent vague, abstract relationships.}
\label{tab:edge_scores}
\resizebox{0.5\columnwidth}{!}{%
\begin{tabular}{lcl}
\toprule
\textbf{Edge Type} & \textbf{Score} & \textbf{Example Edge Connection} \\
\midrule
\makecell[l]{Top-Scored Edges\\(Visual Cue $\rightarrow$ Implication)} & 
\cellcolor{LightGreen}\makecell{3.51\% \\ 3.45\% \\ 3.38\%} & 
\makecell[l]{\texttt{weapon} $\rightarrow$ \texttt{armed robbery} \\ \texttt{mask} $\rightarrow$ \texttt{identity concealed} \\ \texttt{forced entry} $\rightarrow$ \texttt{property crime}} \\
\midrule
\makecell[l]{Bottom-Scored Edges\\(Abstract $\rightarrow$ Abstract)} & 
\cellcolor{LightRed}\makecell{1.88\% \\ 1.81\% \\ 1.75\%} & 
\makecell[l]{\texttt{danger} $\rightarrow$ \texttt{high stress} \\ \texttt{illegal} $\rightarrow$ \texttt{criminal intent} \\ \texttt{theft} $\rightarrow$ \texttt{financial loss}} \\
\bottomrule
\end{tabular}%
}
\end{table}

\paragraph{Pruning of Abstract and General Concepts.}
As detailed in \autoref{tab:node_analysis}, the refinement process demonstrates a clear pattern of shifting from general or abstract concepts to concrete, observable ones that are more relevant to CCTV footage. For the \textbf{Robbery} class, the model prunes general terms like \texttt{theft} and \texttt{danger}, which are high-level descriptions, in favor of specific, visually verifiable objects and actions such as \texttt{weapon}, \texttt{mask}, and \texttt{forced entry}. These retained nodes represent the key visual components that define a robbery in surveillance video. A similar trend is observed for the \textbf{Vandalism} class. The model discards generic descriptors like \texttt{damage} and \texttt{destruction} and instead learns to focus on the tangible evidence of the act, such as \texttt{graffiti}, \texttt{breaking windows}, and \texttt{spray paint}. This shows that \MissionHD\ successfully adapts the reasoning graph to focus on the most salient visual cues in the downstream dataset, effectively grounding the graph in the visual domain.

\paragraph{Strengthening Visually-Grounded Edges.}
Beyond pruning nodes, \MissionHD\ refines the relationships \textit{between} concepts. \autoref{tab:edge_scores} shows that for the \textbf{Robbery} class, the highest-scoring connections are those that link a concrete, observable cue to its direct, logical implication. For instance, the edge \texttt{weapon} $\rightarrow$ \texttt{armed robbery} receives a high contribution score, indicating the model has learned this critical visual distinction. Conversely, edges connecting two abstract concepts, such as \texttt{danger} $\rightarrow$ \texttt{high stress}, are assigned low scores and implicitly pruned, as they lack specific, verifiable visual grounding.

\end{document}

%% file: Sections/0_Abstract.tex
\begin{abstract}
LLM-generated reasoning graphs, referred to as \emph{mission-specific graphs} (MSGs), are increasingly used for video anomaly detection (VAD) and recognition (VAR). However, they are typically treated as fixed despite being generic and distribution-deficient. Conventional graph structure refinement (GSR) methods are ill-suited to this setting, as they rely on learning structural distributions that are absent in LLM-generated graphs. We propose \emph{HDC-constrained Graph Structure Refinement (HDC-GSR)}, a new paradigm that directly optimizes a \emph{decodable}, task-aligned graph representation in a single hyperdimensional space without distribution modeling. Leveraging Hyperdimensional Computing (HDC), our framework encodes graphs via binding and bundling operations, aligns the resulting graph code with downstream loss, and decodes edge contributions to refine the structure. We instantiate this approach as \MissionHD\ for weakly supervised VAD/VAR and demonstrate consistent performance gains on benchmark datasets.
\end{abstract}

%% file: Sections/1_Introduction.tex
Detecting anomalies in human activities through artificial vision is essential for enabling rapid and reliable incident response. To make such systems practical at scale, researchers have focused on weakly supervised video anomaly detection and recognition (VAD/VAR), which mitigates the high cost of frame-level annotations while maintaining competitive performance~\cite{anomalyclip, vad1_cvpr, rtfm, vad2_cvpr, yun2025missiongnn}. Beyond reducing annotation overhead, recent research has also shifted toward mitigating the substantial computational cost of heavy backbone models, particularly large language models (LLMs), which offer strong anomaly detection capabilities but incur significant overhead. A growing body of work addresses this challenge by leveraging \emph{LLM-generated mission-specific graphs} (MSGs) to scaffold structured reasoning~\cite{luo2024rog,qiu2025step,li-etal-2025-graphotter,yun2025missiongnn}. By distilling high-level task knowledge into compact graph structures, these approaches enable interpretable intermediate representations while avoiding repeated, costly end-to-end LLM inference or fine-tuning.
\
\begin{figure}[!t]
    \centering
    \includegraphics[width=0.5\columnwidth]{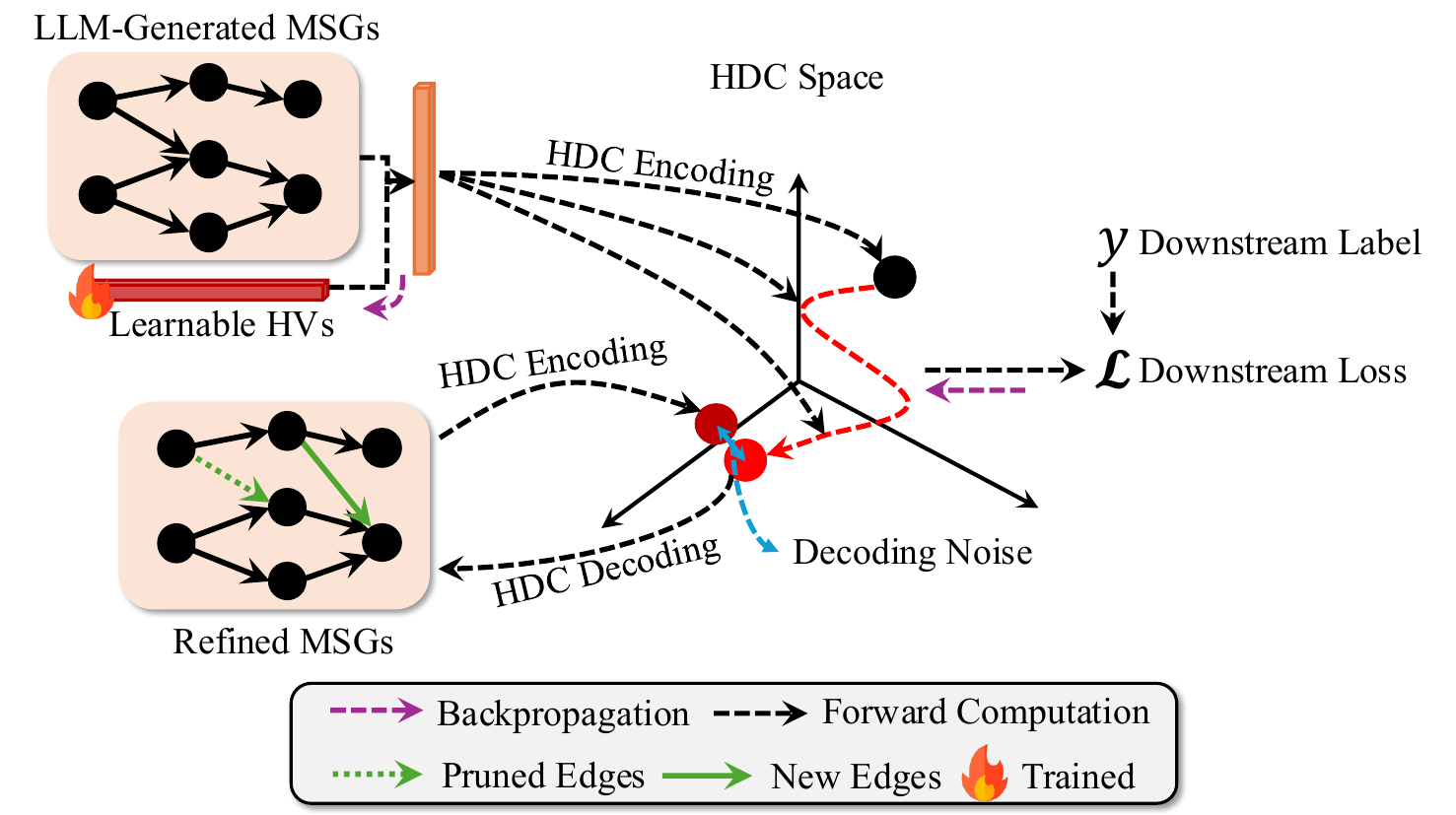}
    \caption{The proposed \MissionHD\ framework refines LLM-generated MSGs through the HDC-GSR paradigm. By leveraging hyperdimensional encoding and decoding, it aligns graph structures with downstream tasks without relying on direct structural-distribution learning.}
    \label{fig:missionhd_pipeline}
    \vspace{-4mm}
\end{figure}
\
Despite their promise, existing approaches typically treat the generated MSG as fixed, adopting it without further adaptation and implicitly assuming that the LLM-produced structure is well aligned with downstream objectives. In practice, such graphs are often generic and may contain irrelevant nodes or edges, which can degrade task performance. This naturally raises the need to refine LLM-generated MSGs so that their structure better supports downstream objectives.

Existing works on Graph Structure Refinement (GSR)~\cite{gsr_23,ecl-gsr} typically learn a graph \emph{structural distribution}, either by refining the given graph itself or by leveraging large-scale graph datasets that share the same distribution as the target graphs. However, LLM-generated graphs for downstream tasks are not well-suited to such approaches for two main reasons. \textbf{First, they lack a meaningful structural distribution.} Since LLM-generated graphs are not observed from real-world domains (e.g., social or citation networks), it is unclear whether they exhibit stable or learnable structural regularities. Moreover, LLMs may produce graphs that are difficult to model structurally, such as graphs that are too small or that exhibit highly skewed connectivity (e.g., tens of thousands of input nodes representing video frames feeding into only a small set of reasoning nodes). \textbf{Second, LLM-generated MSGs are \emph{novel artifacts} that lack large-scale datasets sharing similar structural distributions.} Unlike graphs derived from established domains, there are no large-scale corpora of LLM-generated graphs available for pre-training, rendering conventional GSR methods impractical.
\
To address these limitations—stemming from the difficulty of learning structural distributions and the novel nature of MSGs—we identify a fundamental shortcoming in prior GSR frameworks: they focus on learning graph distributions directly, rather than learning a task-aligned \emph{graph representation}. In the context of LLM-generated MSGs, where structural distributions are weak or nonexistent, it is more appropriate to learn representations that are directly aligned with downstream task objectives, instead of relying on distribution-level modeling.

Motivated by this, we seek an alternative paradigm that avoids structural-distribution learning and directly optimizes task-aligned graph representations. Recent work shows that Hyperdimensional Computing (HDC) provides strong \emph{encode--decode} capabilities for structured data~\cite{kleyko2022survey,poduval2022graphd,dalvi2025hyperdimensional}. Operating in a single high-dimensional space, HDC composes information through simple algebraic operations—\emph{binding} (element-wise multiplication) and \emph{bundling} (element-wise addition)—to produce compact and decodable representations that are robust to noise and do not require modeling an explicit structural distribution. 
\
These properties suggest a principled path toward \emph{task-aligned, decodable} graph optimization: instead of learning a graph distribution, we encode the graph into a hyperdimensional representation, align this representation directly with the downstream task loss, and subsequently decode the optimized representation back into an explicit graph structure. In this way, graph refinement is driven by task supervision rather than by assumptions about an underlying structural distribution.

We formalize this idea as \emph{HDC-constrained Graph Structure Refinement (HDC-GSR)}, a general framework that directly optimizes a \emph{decodable} graph representation in a single hyperdimensional space under downstream task supervision. In HDC-GSR, HDC operations induce a \emph{constrained GNN encoder}: layer hypervectors function as \emph{diagonal} linear maps in the shared space (i.e., an MLP with diagonal weights), \emph{binding} implements message passing, and \emph{bundling} performs aggregation. The resulting graph code is aligned with the task loss, and edge contributions are then \emph{decoded} via HDC similarity probes to update the explicit graph structure. Through this encode–align–decode process, graph refinement is guided directly by task signals, eliminating the need for structural-distribution modeling. Building on this paradigm, we instantiate HDC-GSR for weakly supervised VAD/VAR in \MissionHD, which efficiently encodes hierarchical reasoning paths into compositional hypervectors and decodes edge scores for task-driven graph editing. Replacing the original LLM-generated MSGs with \MissionHD-refined graphs yields consistent performance gains on video anomaly tasks.

Our key contributions are summarized as follows:
\begin{itemize}
    \item We propose \textbf{HDC-constrained Graph Structure Refinement (HDC-GSR)}, a new paradigm for refining LLM-generated \emph{mission-specific graphs} that directly optimizes a \emph{decodable}, task-aligned graph representation without relying on structural-distribution learning.
    
    \item We design a compact \emph{encode--align--decode} framework in a single hyperdimensional space, where binding and bundling induce a \emph{constrained GNN} (with diagonal layer maps), and edge refinement is achieved through compositional similarity-based decoding.
    
    \item We instantiate this paradigm as \MissionHD\ for weakly supervised VAD/VAR and demonstrate consistent improvements over strong baselines on benchmark datasets.
\end{itemize}

%% file: Sections/2_Preliminaries.tex
\subsection{Hyperdimensional Computing}

HDC is a vector-symbolic computational approach inspired by human cognitive processes that encodes concepts as high-dimensional vectors, typically exceeding $1,000$ dimensions~\cite{kanerva2009hyperdimensional, thomas2021theoretical}. Leveraging HDC-specific mathematical operations—namely bundling and binding—this framework enables a range of machine learning tasks such as classification, regression, and link prediction across diverse modalities, including graphs~\cite{poduval2022graphd, chen2024hdreason, jeong2025exploiting}. In this section, we provide a concise overview of the fundamental operations that form the foundation of our proposed method.

Let $\mathcal{H}=\mathbb{R}^D$ be the $D$-dimensional hypervector (HV) space. $D$ commonly ranging from thousands to hundreds of thousands. A hypervector $\mathbf{h}\in\mathcal{H}$ is compared via the normalized dot product 
$
\delta(\mathbf{h}_1,\mathbf{h}_2)=\frac{1}{D}\,\mathbf{h}_1^{\!\top}\mathbf{h}_2.
$
Three algebraic operations endow the space with symbolic reasoning capability:

\begin{description}
  \item[Bundling] $\mathbf{h}=\bigoplus_{i=1}^k\mathbf{h}_i$ (component-wise addition) is a \emph{set} operation that preserves similarity: $\delta(\mathbf{h},\mathbf{h}_i) > 0$.
  \item[Binding]  $\mathbf{h}=\mathbf{h}_1\otimes\mathbf{h}_2$ (component-wise multiplication) is an \emph{association} yielding a vector nearly orthogonal to its factors: $\delta(\mathbf{h},\mathbf{h}_1) \approx 0$. Binding is similarity-preserving under a third operand:  
        $\delta(\mathbf{v}\otimes\mathbf{h}_1,\mathbf{v}\otimes\mathbf{h}_2) \simeq \delta(\mathbf{h}_1,\mathbf{h}_2)$.
  \item[Permutation] $\rho(\mathbf{h})$ applies a fixed random permutation; $\rho^p$ produces $p$-step positional encodings and is invertible, $\rho^{-p}(\rho^p(\mathbf{h}))=\mathbf{h}$.
\end{description}

\paragraph{Graph encoding/decoding in HDC}
In prior HDC approaches for general graph structures, such as GrapHD~\cite{poduval2022graphd}, an unweighted graph $G = (\mathcal{V}, \mathcal{E})$ is encoded using a composition of node and neighborhood information. Each node $v_i \in \mathcal{V}$ is assigned a random base hypervector $\mathbf{E}_i \in \mathcal{H}$. Its local neighborhood is captured by a memory vector $\mathbf{M}_i = \bigoplus_{v_j \in \mathcal{N}(i)} \mathbf{E}_j$, where $\mathcal{N}(i)$ denotes the set of adjacent nodes. The overall graph hypervector is then computed as:
$$
\mathbf{E}_G = \frac{1}{|\mathcal{V}|} \sum_{i} \mathbf{E}_i \otimes \mathbf{M}_i.
$$
This representation allows for graph-level comparisons using similarity functions and enables edge decoding through unbinding operations and thresholding on similarity scores. Extensions of this method support directed graphs via fixed permutations and weighted graphs via stochastic encoding.
\
While this framework is general and efficient, it does not explicitly capture sequential or hierarchical semantics often required in decision reasoning tasks. In this work, we introduce a specialized hyperdimensional encoding method tailored for reasoning path-based structures such as mission-specific reasoning graphs.

\subsection{Visual Representation Learning Using Graphs}

A graph is a powerful data structure for capturing relationships between entities. The application of graphs to solve vision tasks has been widely studied, ranging from obtaining scene graphs to represent connections between objects in a single image~\cite{xu2017scenegraph, huang2025building}, to organizing both verbal and visual clues for question answering~\cite{vqa-gnn}, and to video anomaly detection~\cite{zhong2019graph_mil_video, xdv, masuakwa2025pv}. In particular, novel methods that facilitate the decision-making process of machine-learning models with the aid of knowledge graphs have proven effective across various tasks~\cite{luo2024rog,qiu2025step,li-etal-2025-graphotter}. Among these, MissionGNN~\cite{yun2025missiongnn} has shown promising results in video anomaly detection. They construct mission-specific graphs (MSGs) layered directed acyclic graphs (DAGs) $G = (\mathcal{V}, \mathcal{E})$ constructed dynamically from task instructions, external knowledge bases. Nodes $v_k^{(i)} \in \mathcal{V}$ are assigned to semantic levels $i \in \{1, \dots, \ell\}$, and edges are only allowed between adjacent layers, i.e., $(v_k^{(i)}, v_t^{(i+1)}) \in \mathcal{E}$, enforcing a level-monotonic hierarchy.
\
Each node encodes a task-relevant concept such as object attributes, spatial relations, or symbolic goals, and is associated with a feature vector $\mathbf{x}^{(i)}_k \in \mathbb{R}^m$. Reasoning in MSGs proceeds along reasoning paths---maximal sequences $(v_{k_1}^{(1)}, \dots, v_{k_\ell}^{(\ell)})$ tracing causal chains from sensor-level observations to high-level decisions~\cite{masuakwa2025pv}. For example, a path might encode: ``mug on table'' $\rightarrow$ ``is dirty'' $\rightarrow$ ``move to sink''.

However, a core limitation in the MSG-based approaches is that their generated MSG nodes and edges reflect what LLMs  ``see,'' rather than what the visual backbones and subsequent decision-making layers of the VAD model observe. Recent research indicates that LLMs can indeed imagine visual worlds ~\cite{sharma2024vision}; nonetheless, there remains a significant discrepancy between actual visual data and the visual clues inferred by LLMs, as these models are pre-trained predominantly on extensive text corpora. Motivated by this limitation, we propose \MissionHD, a novel framework that dynamically reformulates MSGs during each training epoch, aligning them more closely with the representations perceived by the final decision-making layer, aided by HDC.

%% file: Sections/3_Methodology.tex
\begin{figure}[!t]
    \centering
    \includegraphics[width=0.5\columnwidth]{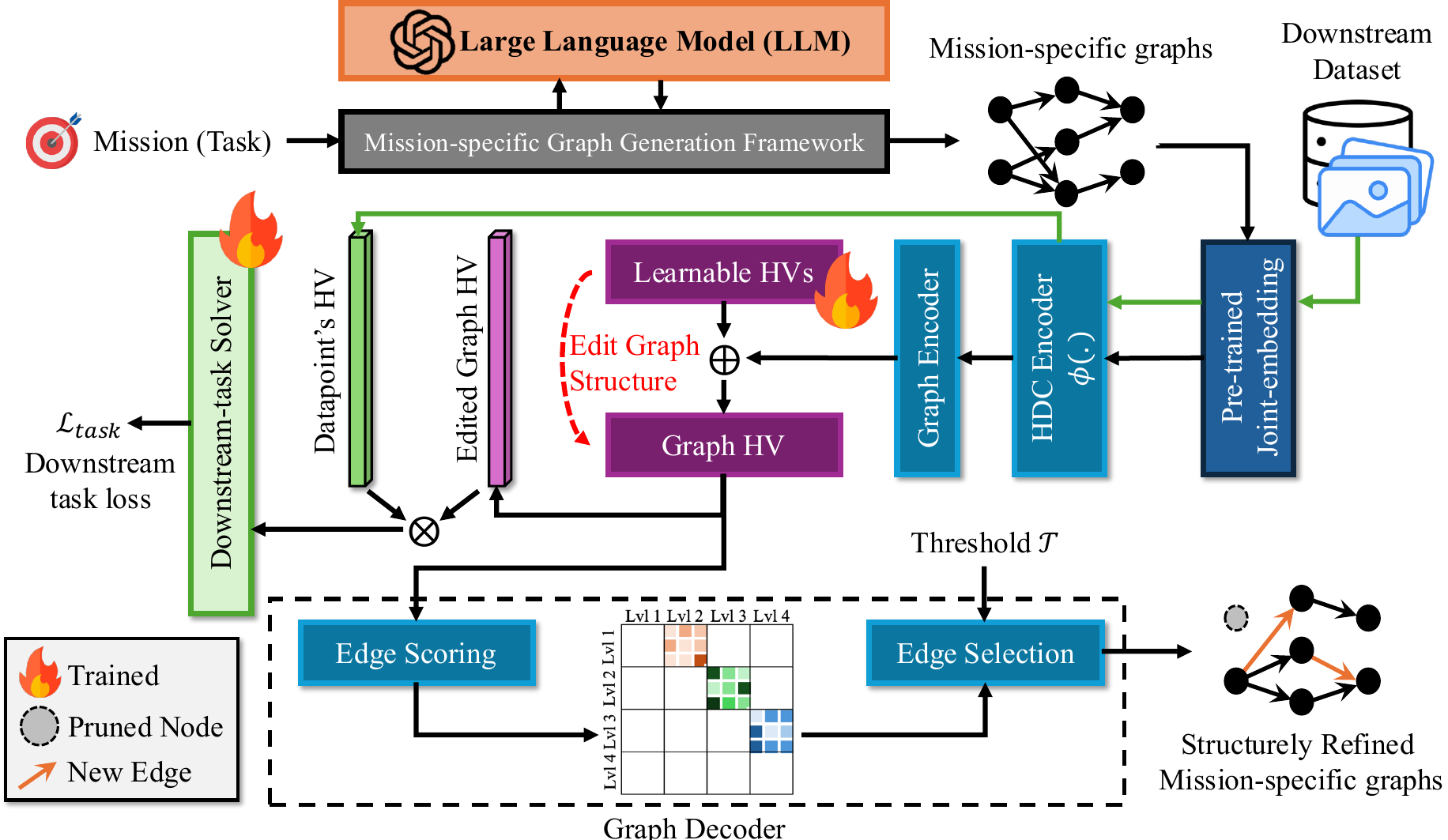}
    \caption{Overview of the \MissionHD\ pipeline. We propose a hyperdimensional encoding and refinement framework for mission-specific reasoning graphs, enabling efficient symbolic representation and task-driven structure updates via edge contribution scoring.}
    \label{fig:missionhd_pipeline}
    \vspace{-4mm}
\end{figure}

Our framework, \MissionHD, refines LLM-generated reasoning graphs by learning their optimal structure for downstream visual tasks. As illustrated in \autoref{fig:missionhd_pipeline}, the core of our approach is an end-to-end trainable framework that leverages HDC to create a task-aligned, decodable representation of the graph. We first encode the graph's structure and semantics into a single hypervector, which is then optimized via a downstream task loss that implicitly learns structural edits. Finally, we decode this optimized hypervector back into a refined graph structure by scoring and selecting the most salient edges.

\subsection{Setup and Notation}
Let $G=(V,E)$ be a layered Directed Acyclic Graph (DAG) with layers $\{V^{(i)}\}_{i=1}^{\ell}$. Each node $v^{(i)}_k \in V^{(i)}$ has an associated feature vector $x^{(i)}_k\in\mathbb{R}^m$. We use $\phi:\mathbb{R}^m\to\mathbb{R}^D$ as a projection into the $D$-dimensional HDC space, yielding node hypervectors (HVs) $H^{(i)}_k=\phi(x^{(i)}_k)$ and layer HVs $L^{(i)}$. The fundamental HDC operations are element-wise binding ($\otimes$), bundling ($\oplus$), and cosine similarity ($\delta(a,b)=\frac{a^\top b}{\|a\| \|b\|}$). For simplicity and clarity of notation, the sensor and final encoding nodes are omitted from the formulas.

\subsection{Hyperdimensional Graph Encoding and Optimization}
Our method centers on creating and optimizing a single, holistic hypervector representation of the reasoning graph that is aligned with the downstream task.

\begin{figure}[!t]
    \centering
    \includegraphics[width=0.5\columnwidth]{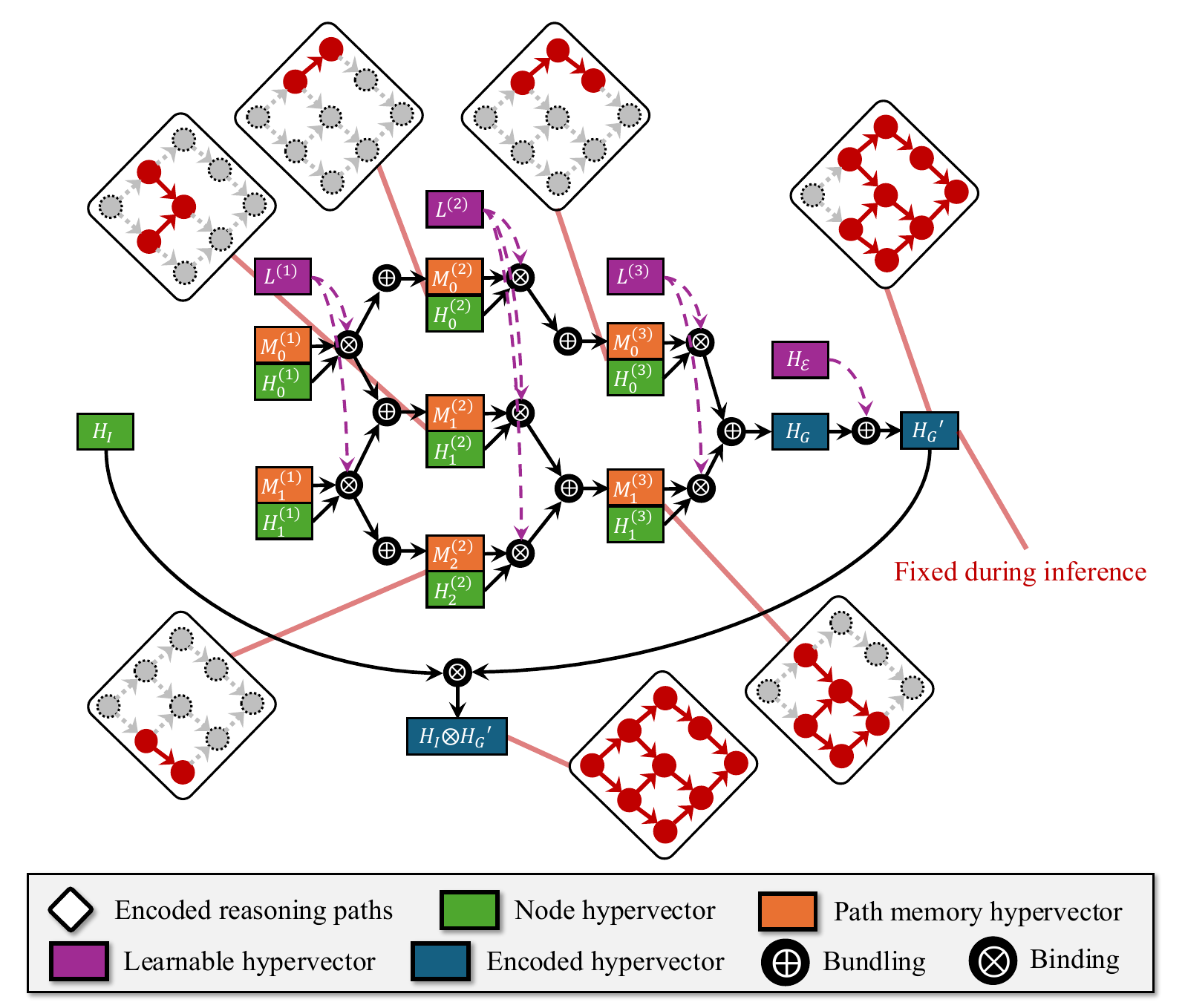}
    \caption{Overview of the \MissionHD\ encoding pipeline. A layered DAG has its node features projected into high-dimensional space. A DP approach then efficiently computes the global graph hypervector. The final hypervector is augmented with learnable structural edits and fused with sensor features for downstream decision-making.}
    \label{fig:MissionHD_overview}
    \vspace{-4mm}
\end{figure}

\paragraph{MSG Encoding.} The encoding pipeline is depicted in \autoref{fig:MissionHD_overview}. To create a computationally tractable representation, we encode all reasoning paths into a graph hypervector, $H_G$, using a Dynamic Programming approach. We compute \textit{path memory hypervectors} $M_k^{(i)}$ that recursively bundle all paths leading to a node $v_k^{(i)}$:
\begin{equation}
    M_k^{(i)}=\bigoplus_{v_t^{(i-1)}\rightarrow v_k^{(i)}}L^{(i-1)}\otimes M_t^{(i-1)}\otimes H_t^{(i-1)}
\end{equation}
The final graph hypervector $H_G$ is the bundle of all paths reaching the last layer: $H_G=\bigoplus_{v_k^{(\ell)}}L^{(\ell)}\otimes M_k^{(\ell)}\otimes H_k^{(\ell)}$ with linear time complexity $\mathcal{O}(|V|+|E|)$. The full encoding procedure is detailed in \autoref{alg:encoding}.

\paragraph{Task-Driven Structure Optimization.} To enable refinement, we introduce a learnable \textit{structural edit hypervector}, $H_{\mathcal{E}}$. This vector is generated by projecting a low-dimensional latent vector $z_{\mathcal{E}} \in \mathbb{R}^d$ via a learnable matrix $P \in \mathbb{R}^{D \times d}$:
\begin{equation}
    H_{\mathcal{E}} = P z_{\mathcal{E}}
\end{equation}
This vector is bundled with the static graph encoding to form an augmented, trainable graph hypervector, $H'_G = H_G \oplus H_{\mathcal{E}}$. Given a sensor input $I$, the final input provided to the decision model is $H_{\text{input}} = \phi(I) \oplus H'_G$. The model's parameters $\theta$ (which include $P$ and the decision model weights) are optimized by minimizing a composite loss function over the data distribution $\mathcal{D}$:
\begin{equation}
    \theta^* = \arg\min_{\theta} \mathbb{E}_{(I, y) \sim \mathcal{D}}[\mathcal{L}_{\text{task}}(f_{\theta}(I, G), y)]
\end{equation}
where $\mathcal{L}_{\text{task}} = \mathcal{L}_{\text{CE}} + \lambda \mathcal{L}_{\text{smooth}}$. The temporal smoothness regularizer penalizes large changes in anomaly predictions $a_t$ for consecutive video frames: $\mathcal{L}_{\text{smooth}} = \frac{1}{T-1}\sum_{t=2}^{T} (a_t - a_{t-1})^2$. This task-driven process optimizes $H_{\mathcal{E}}$ to represent the edits that align the graph's structure with the task.

\subsection{Graph Structure Refinement via Hyperdimensional Decoding}
After training, the optimized graph vector $H'_G$ is used to decode a refined graph structure. This process begins by computing forward ($F_k^{(i)}$) and backward ($B_k^{(i)}$) context vectors under a fully-connected assumption between layers to evaluate all candidate edges:
\begin{align}
    F_k^{(i)} &= \bigoplus_{v_t^{(i-1)} \in V^{(i-1)}} \mathrm{Norm}(L^{(i-1)} \otimes F_t^{(i-1)} \otimes H_t^{(i-1)}) \\
    B_k^{(i)} &= \bigoplus_{v_t^{(i+1)} \in V^{(i+1)}} \mathrm{Norm}(L^{(i+1)} \otimes B_t^{(i+1)} \otimes H_t^{(i+1)})
\end{align}
For each candidate edge $(v_k^{(i)} \rightarrow v_t^{(i+1)})$, we construct a compositional probe hypervector, $\mathrm{FB}(i,k,t) = L^{(i)} \otimes F_k^{(i)} \otimes H_k^{(i)} \otimes L^{(i+1)} \otimes B_t^{(i+1)} \otimes H_t^{(i+1)}$. Its contribution score is computed via softmax normalization of its similarity to the target vector:
\begin{equation}\label{eq:edgescore}
    s_{k,t}^{(i)} = \frac{\exp(\delta(\mathrm{FB}(i,k,t), H'_G))}{\sum_{t' \in V^{(i+1)}} \exp(\delta(\mathrm{FB}(i,k,t'), H'_G))}
\end{equation}
The refined edge set $\mathcal{E}_{\text{refined}}$ is then formed by applying a threshold $\mathcal{T}$:
\begin{equation}
    \mathcal{E}_{\text{refined}} = \{ (v_k^{(i)}, v_t^{(i+1)}) \mid s_{k,t}^{(i)} > \mathcal{T} \}
\end{equation}
This procedure is detailed in \autoref{alg:denseencoding} and \autoref{alg:decoding}.

\begin{figure*}[!t]
    \centering
    \includegraphics[width=\textwidth]{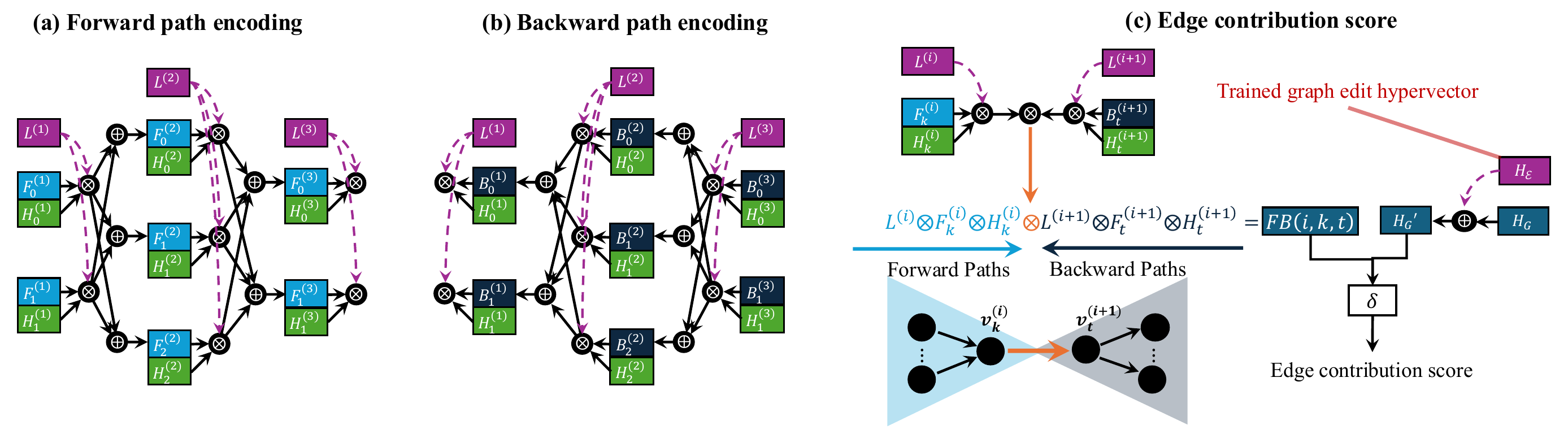}
    \caption{Illustration of the hyperdimensional graph structure refinement in \MissionHD. (a) Forward path encodings are computed assuming full connectivity. (b) Backward path encodings are computed in reverse. (c) A hypothetical hypervector for a candidate edge is constructed and compared against the trained graph vector to compute an edge contribution score.}
    \label{fig:MissionHD_edge}
    \vspace{-4mm}
\end{figure*}

\section{Theoretical Analysis}
We ground our framework in GNN theory and derive bounds on decoding accuracy, providing theoretical justification for the proposed encoding and decoding scheme. Complete details and proofs are given in \autoref{appx:thereticalanalysis}.

\paragraph{HDC-Constrained GNN Equivalence.}
\begin{theorem}[HDC layer $\equiv$ diagonal linear map]
\label{thm:hdcgnneq}
An HDC layer operation, $z \mapsto L^{(i)}\otimes z$, is mathematically equivalent to a linear layer with a diagonal weight matrix $D_{L^{(i)}}:=\mathrm{diag}(L^{(i)})$.
\end{theorem}

\begin{theorem}[\MissionHD\ as a path-sum GNN]\label{trm:pathsumgnn}
\label{thm:pathsumgnn}
A hierarchical GNN with diagonal linear maps per layer ($W^{(i)}=D_{L^{(i)}}$) and element-wise message functions (realized by binding) produces a graph-level representation equal to the HDC path-sum representation $H_G$.
\end{theorem}

\paragraph{Edge score consistency.}
Recall that for an edge $(v^{(i)}_k\!\to v^{(i+1)}_t)$ we form the probe
$FB(i,k,t)$ (\autoref{fig:MissionHD_edge}) and score it against the trained target vector $H'_G = H_G \oplus H_\mathcal{E}$.
Let the cosine score be $S_{i,k,t} := \delta\!\big(FB(i,k,t), H'_G\big)$, and denote by $P_{k\to t}$ the set of paths that contain $(k\!\to\! t)$ and by $P_{\neg(k\to t)}$ those that do not.
The encoder bundles all path hypervectors, so $H_G$ is (up to the normalization done in \autoref{alg:encoding}) the bundle $\bigoplus_{p\in\mathcal{P}} H_{\text{path}}(p)$ with $\|H_{\text{path}}(p)\|_2=1$.

\begin{proposition}[Edge score consistency]
\label{prop:attr-consistency-strong}
Let $\widehat{H}'_G:= H'_G/\|H'_G\|_2$ and $\widehat{H}_\mathcal{E}:= H_\mathcal{E}/\|H'_G\|_2$.
Then $S_{i,k,t} = \langle FB(i,k,t), \widehat{H}'_G\rangle$ admits the decomposition
\begin{align*}
S_{i,k,t}
\,&=\, \underbrace{\sum_{p\in P_{k\to t}}\!\!\delta\!\big(FB(i,k,t), H_{\text{path}}(p)\big)}_{\text{signal from paths containing }(k\!\to\! t)}
\\\;&+\;
\underbrace{\delta\!\big(FB(i,k,t), \widehat{H}_\mathcal{E}\big)}_{\text{projection of edit}}
\;+\;
\xi_{i,k,t},
\end{align*}
where the residual aggregates the non-containing paths,
\[
\xi_{i,k,t} \;=\; \sum_{p\in P_{\neg (k\to t)}} \delta\!\big(FB(i,k,t), H_{\text{path}}(p)\big).
\]
Under Assumptions~\ref{assump:assumptions}, each inner product $\delta(FB(i,k,t), H_{\text{path}}(p))$ with $p\in P_{\neg (k\to t)}$ is an average of products of i.i.d.\ coordinates between (approximately) independent, near-orthogonal random directions in $\mathbb{R}^D$; thus it has mean $0$ and sub-Gaussian tails
$\Pr(|Z|>\varepsilon)\le 2\exp(-cD\varepsilon^2)$ for some absolute $c>0$ (Hoeffding-type concentration).
Sums of $n$ independent/weakly-dependent sub-Gaussian terms remain sub-Gaussian with parameter scaling linearly in $n$, hence $\xi_{i,k,t}$ is sub-Gaussian with parameter
\[
\sigma_{i,k,t}^2 \;=\; \frac{c}{D}\,\big|P_{\neg(k\to t)}\big|.
\]
Consequently, for any $\varepsilon>0$,
\begin{align*}
\Pr\!\Big(\big|S_{i,k,t}-\mu_{i,k,t}\big| \ge \varepsilon\Big)
\;&\le\;
2\exp\!\left(-\,\frac{\varepsilon^2}{2\sigma_{i,k,t}^2}\right)
\\\;&=\;
2\exp\!\left(-\,\frac{D\,\varepsilon^2}{2c\,|P_{\neg(k\to t)}|}\right),
\end{align*}
where
\[
\mu_{i,k,t} := \sum_{p\in P_{k\to t}}\!\!\delta\!\big(FB(i,k,t), H_{\text{path}}(p)\big)+\delta\!\big(FB(i,k,t), \widehat{H}_E\big).
\]
\end{proposition}

\paragraph{Decoding guarantees.}
\begin{proposition}[Decoding accuracy bound]
\label{prop:decode-accuracy-strong}
Let $E_{\text{refined}}(\mathcal{T})$ be the edges kept by \autoref{alg:decoding} at threshold $\mathcal{T}\in(0,1)$, and let $\mathcal{P}_{\text{keep}}(\mathcal{T})$ and $\mathcal{P}_{\text{miss}}(\mathcal{T})$ denote, respectively, the original paths whose every edge lies in $E_{\text{refined}}(\mathcal{T})$ and those that do not.
Let $Q$ be a finite set of newly synthesized paths added to approximate the learned edit vector $H_\mathcal{E}$ (which is not produced by \autoref{alg:encoding}).
Let $H_G^{\text{refined}}$ be the re-encoded hypervector from the refined graph using $\mathcal{P}_{\text{keep}}(\mathcal{T})\cup Q$, and write $\widehat{H}'_G := H'_G/\|H'_G\|_2$.
Then, under Assumptions~\ref{assump:assumptions},
\begin{align*}
1-\big(\delta(H_G^{\text{refined}}, H'_G)\big)^2
\;=\;
O\!\Big(\underbrace{\sum_{p\in \mathcal{P}_{\text{miss}}(\mathcal{T})} \big(\tau_{\max}(\mathcal{T})\big)^{2L(p)}}_{\text{discarded originals}}\Big)
\\\;+\;
O\!\Big(\underbrace{\big\|\,\widehat{H}_\mathcal{E} - \Pi_{\mathrm{span}(\{H_{\text{path}}(q):\,q\in Q\})}\widehat{H}_\mathcal{E}\,\big\|_2^2}_{\text{new-path approximation}}\;\\+\;\underbrace{\tfrac{|Q|}{D}}_{\text{finite-$D$ crosstalk}}\Big)
\;+\;
O\!\Big(\underbrace{\tfrac{|\mathcal{P}_{\text{keep}}(\mathcal{T})|}{D}}_{\text{finite-$D$ crosstalk}}\Big),
\end{align*}
where $L(p)$ is the number of edges in path $p$, $\tau_{\max}(\mathcal{T})\in(0,1)$ is a threshold–dependent damping factor estimated from observed softmax margins (paths with at least one edge below $\mathcal{T}$ are damped), and $\Pi$ denotes the orthogonal projector onto the indicated span. In particular, for approximately uniform path length $L$ and $K_{\text{miss}}(\mathcal{T}):=|\mathcal{P}_{\text{miss}}(\mathcal{T})|$,
\begin{align*}
1-\big(\delta(H_G^{\text{refined}}, H'_G)\big)^2
\;=\;
O\!\Big(K_{\text{miss}}(\mathcal{T})\,(\tau_{\max}(\mathcal{T}))^{2L}\Big)
\\\;+\;
O\!\Big(\big\|\,\widehat{H}_\mathcal{E} - \Pi_{\mathrm{span}(Q)}\widehat{H}_\mathcal{E}\,\big\|_2^2 + \tfrac{|Q|}{D} + \tfrac{|\mathcal{P}_{\text{keep}}(\mathcal{T})|}{D}\Big).
\end{align*}

\end{proposition}

%% file: Sections/4_Experiments.tex
\begin{figure}[!t]
    \centering
    \includegraphics[width=0.5\columnwidth]{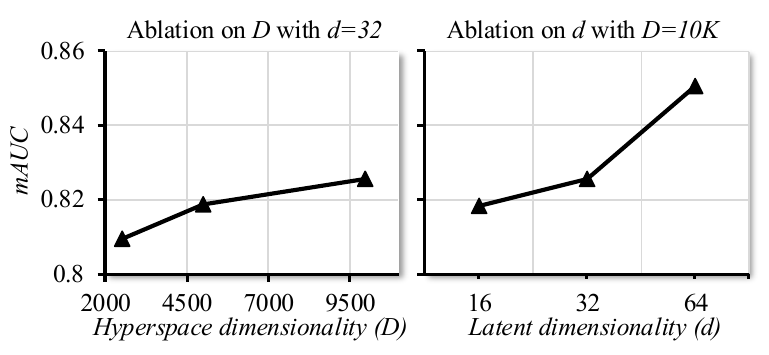}
    \caption{Ablation on two dimensionalities: hyperspace (left) and latent space (right) on UCF-Crime dataset. Increasing latent dimensionality yields larger performance gains than increasing hyperspace dimensionality.
}\label{fig:ablation_dim}
\end{figure}

\begin{figure}[!t]
    \centering
    \includegraphics[width=0.4\columnwidth]{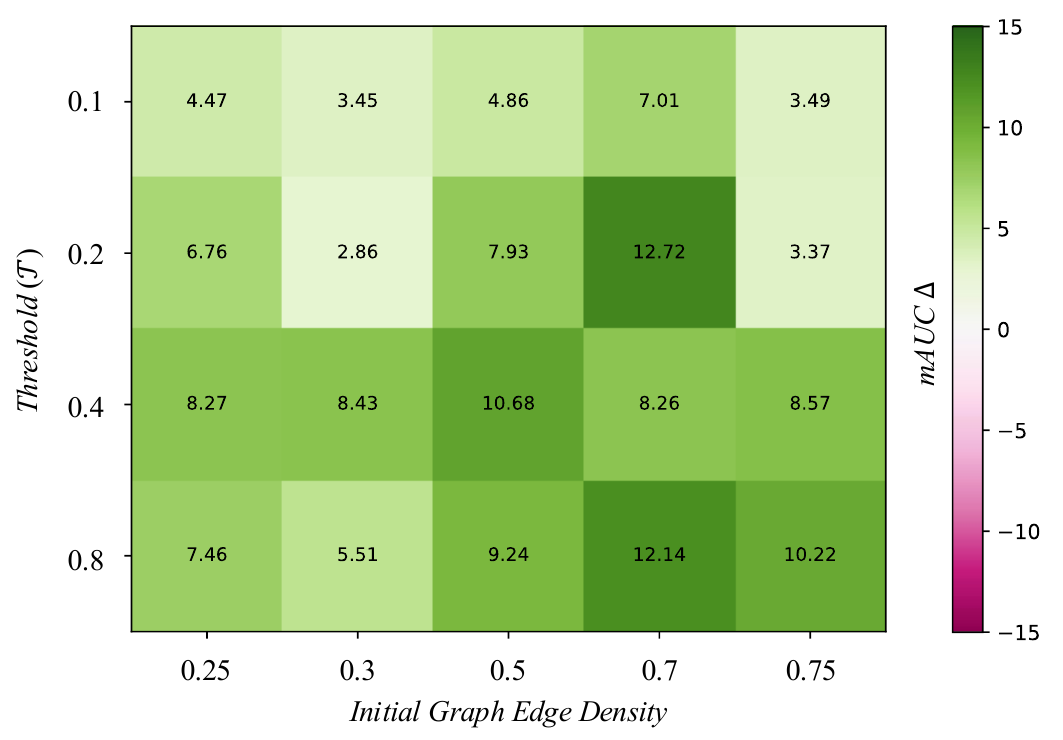}
    \caption{VAR score differences (mAUC $\Delta \uparrow$, \%) under varying initial MSGs and thresholds on UCF-Crime dataset. 
Initial MSGs are randomly augmented according to edge density derived from LLM-generated MSGs. 
Differences are computed between mAUC scores from direct use of randomly augmented MSGs and from our proposed refinement.
}\label{fig:edgedensity_mAUCs}
\end{figure}

\begin{figure*}[!t]
    \centering
    \includegraphics[width=1.0\textwidth]{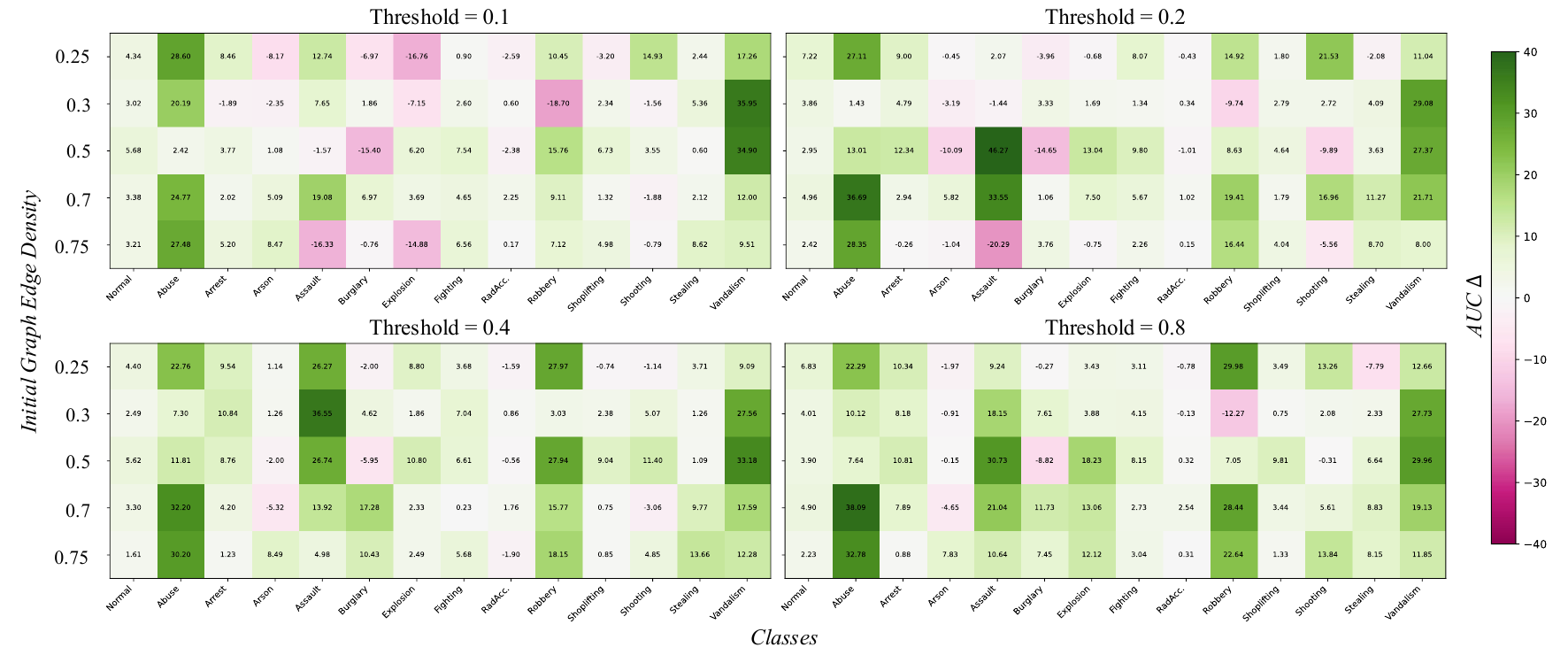}
    \caption{Class-wise VAD score differences (AUC $\Delta \uparrow$, \%) for all anomaly classes under varying initial MSGs and thresholds on UCF-Crime dataset. 
Initial MSGs are randomly augmented according to edge density derived from LLM-generated MSGs. 
Differences are computed between AUC scores from direct use of randomly augmented MSGs and from our proposed refinement.
}\label{fig:edgedensity_AUCs}
\end{figure*}

\begin{table*}[t]
\centering
\caption{
AUC scores (\%) for VAD (Normal) and VAR on the UCF-Crime dataset using different refinement methods. Relative changes from the baseline (None) are shown below each value (green = improvement, red = degradation).
}
\label{tab:vad_var_auc_compact}
\resizebox{\textwidth}{!}{
\begin{tabular}{c|c|ccccccccccccc|c}
\toprule
\textbf{Refine Method} & \textbf{Normal} & \textbf{Abuse} & \textbf{Arrest} & \textbf{Arson} & \textbf{Assault} & \textbf{Burglary} & \textbf{Explosion} & \textbf{Fighting} & \textbf{RoadAcc.} & \textbf{Robbery} & \textbf{Shoplifting} & \textbf{Shooting} & \textbf{Stealing} & \textbf{Vandalism} & \textbf{mAUC} \\
\midrule
None & 
92.14 & 65.59 & 81.47 & \textbf{89.45} & 27.42 & 52.30 & 72.67 & 90.48 & 95.11 & 47.64 & 92.88 & 67.58 & 88.60 & 76.58 & 72.91 \\
\midrule
Adjacency matrix &
\makecell{90.52\\\cellcolor{LightRed}-1.62} & \makecell{79.31\\\cellcolor{LightGreen}+13.72} & \makecell{81.49\\\cellcolor{LightGreen}+0.02} & \makecell{86.58\\\cellcolor{LightRed}-2.87} & \makecell{41.32\\\cellcolor{LightGreen}+13.9} & \makecell{63.90\\\cellcolor{LightGreen}+11.6} & \makecell{72.79\\\cellcolor{LightGreen}+0.12} & \makecell{93.85\\\cellcolor{LightGreen}+3.37} & \makecell{97.37\\\cellcolor{LightGreen}+2.26} & \makecell{56.94\\\cellcolor{LightGreen}+9.3} & \makecell{\textbf{94.43}\\\cellcolor{LightGreen}\textbf{+1.55}} & \makecell{70.87\\\cellcolor{LightGreen}+3.29} & \makecell{80.67\\\cellcolor{LightRed}-7.93} & \makecell{82.72\\\cellcolor{LightGreen}+6.14} & \makecell{77.09\\\cellcolor{LightGreen}+4.19} \\
\midrule
GCN &
\makecell{93.60\\\cellcolor{LightGreen}+1.46} & \makecell{65.55\\\cellcolor{LightRed}-0.04} & \makecell{79.04\\\cellcolor{LightRed}-2.43} & \makecell{84.15\\\cellcolor{LightRed}-5.3} & \makecell{51.69\\\cellcolor{LightGreen}+24.27} & \makecell{77.37\\\cellcolor{LightGreen}+25.07} & \makecell{\textbf{85.91}\\\cellcolor{LightGreen}\textbf{+13.24}} & \makecell{90.59\\\cellcolor{LightGreen}+0.11} & \makecell{97.09\\\cellcolor{LightGreen}+1.98} & \makecell{\textbf{65.58}\\\cellcolor{LightGreen}\textbf{+17.94}} & \makecell{92.39\\\cellcolor{LightRed}-0.49} & \makecell{71.90\\\cellcolor{LightGreen}+4.32} & \makecell{91.91\\\cellcolor{LightGreen}+3.31} & \makecell{85.50\\\cellcolor{LightGreen}+8.92} & \makecell{79.89\\\cellcolor{LightGreen}+6.99} \\
\midrule
GAT &
\makecell{93.81\\\cellcolor{LightGreen}+1.67} & \makecell{75.27\\\cellcolor{LightGreen}+9.68} & \makecell{81.95\\\cellcolor{LightGreen}+0.48} & \makecell{78.99\\\cellcolor{LightRed}-10.46} & \makecell{33.20\\\cellcolor{LightGreen}+5.78} & \makecell{78.52\\\cellcolor{LightGreen}+26.22} & \makecell{82.19\\\cellcolor{LightGreen}+9.52} & \makecell{94.16\\\cellcolor{LightGreen}+3.68} & \makecell{96.15\\\cellcolor{LightGreen}+1.04} & \makecell{59.23\\\cellcolor{LightGreen}+11.59} & \makecell{90.83\\\cellcolor{LightRed}-2.05} & \makecell{69.47\\\cellcolor{LightGreen}+1.89} & \makecell{92.86\\\cellcolor{LightGreen}+4.26} & \makecell{86.45\\\cellcolor{LightGreen}+9.87} & \makecell{78.40\\\cellcolor{LightGreen}+5.5} \\
\midrule
ECL-GSR &
\makecell{95.11\\\cellcolor{LightGreen}+2.97} & \makecell{71.53\\\cellcolor{LightGreen}+5.94} & \makecell{\textbf{87.77}\\\cellcolor{LightGreen}\textbf{+6.3}} & \makecell{78.02\\\cellcolor{LightRed}-11.43} & \makecell{\textbf{66.98}\\\cellcolor{LightGreen}\textbf{+39.56}} & \makecell{82.04\\\cellcolor{LightGreen}+29.74} & \makecell{79.73\\\cellcolor{LightGreen}+7.06} & \makecell{\textbf{95.41}\\\cellcolor{LightGreen}\textbf{+4.93}} & \makecell{95.26\\\cellcolor{LightGreen}+0.15} & \makecell{64.61\\\cellcolor{LightGreen}+16.97} & \makecell{92.11\\\cellcolor{LightRed}-0.77} & \makecell{58.94\\\cellcolor{LightRed}-8.64} & \makecell{\textbf{95.27}\\\cellcolor{LightGreen}\textbf{+6.67}} & \makecell{81.60\\\cellcolor{LightGreen}+5.02} & \makecell{80.71\\\cellcolor{LightGreen}+7.80} \\
\midrule
\MissionHD\ &
\makecell{\textbf{95.67}\\\cellcolor{LightGreen}\textbf{+3.53}} & \makecell{\textbf{88.30}\\\cellcolor{LightGreen}\textbf{+22.71}} & \makecell{84.63\\\cellcolor{LightGreen}+3.16} & \makecell{89.19\\\cellcolor{LightRed}-0.26} & \makecell{47.09\\\cellcolor{LightGreen}+19.67} & \makecell{\textbf{83.19}\\\cellcolor{LightGreen}\textbf{+30.89}} & \makecell{78.29\\\cellcolor{LightGreen}+5.62} & \makecell{91.42\\\cellcolor{LightGreen}+0.94} & \makecell{\textbf{97.58}\\\cellcolor{LightGreen}\textbf{+2.47}} & \makecell{62.55\\\cellcolor{LightGreen}+14.91} & \makecell{93.24\\\cellcolor{LightGreen}+0.36} & \makecell{\textbf{72.32}\\\cellcolor{LightGreen}\textbf{+4.74}} & \makecell{92.23\\\cellcolor{LightGreen}+3.63} & \makecell{\textbf{93.30}\\\cellcolor{LightGreen}\textbf{+16.72}} & \makecell{\textbf{82.56}\\\cellcolor{LightGreen}\textbf{+9.66}} \\
\bottomrule
\end{tabular}
}
\end{table*}

\begin{table}[t]
\centering
\caption{
AUC scores (\%) for VAD (Normal) and VAR on the XD-Violence dataset using different refinement methods. Relative changes from the baseline (None) are shown below each value (green = improvement, red = degradation).
}
\label{tab:xdv_auc_results}
\resizebox{0.5\textwidth}{!}{
\begin{tabular}{c|c|cccccc|c}
\toprule
\textbf{Refine Method} & \textbf{Normal} & \textbf{Abuse} & \textbf{CarAccident} & \textbf{Explosion} & \textbf{Fighting} & \textbf{Riot} & \textbf{Shooting} & \textbf{mAUC} \\
\midrule
None & 
92.70 & 88.91 & 97.79 & 94.70 & 90.92 & 94.99 & 91.60 & 93.15 \\
\midrule
Adjacency matrix & 
\makecell{92.47\\\cellcolor{LightRed}-0.23} & \makecell{89.52\\\cellcolor{LightGreen}+0.61} & \makecell{98.06\\\cellcolor{LightGreen}+0.27} & \makecell{\textbf{94.84}\\\cellcolor{LightGreen}\textbf{+0.14}} & \makecell{91.37\\\cellcolor{LightGreen}+0.45} & \makecell{94.27\\\cellcolor{LightRed}-0.72} & \makecell{91.21\\\cellcolor{LightRed}-0.39} & \makecell{93.21\\\cellcolor{LightGreen}+0.06} \\
\midrule
GCN & 
\makecell{92.14\\\cellcolor{LightRed}-0.56} & \makecell{89.51\\\cellcolor{LightGreen}+0.6} & \makecell{\textbf{98.88}\\\cellcolor{LightGreen}\textbf{+1.09}} & \makecell{94.21\\\cellcolor{LightRed}-0.49} & \makecell{91.00\\\cellcolor{LightGreen}+0.08} & \makecell{95.50\\\cellcolor{LightGreen}+0.51} & \makecell{91.87\\\cellcolor{LightGreen}+0.27} & \makecell{93.49\\\cellcolor{LightGreen}+0.34} \\
\midrule
GAT & 
\makecell{92.79\\\cellcolor{LightGreen}+0.09} & \makecell{90.49\\\cellcolor{LightGreen}+1.58} & \makecell{97.90\\\cellcolor{LightGreen}+0.11} & \makecell{94.03\\\cellcolor{LightRed}-0.67} & \makecell{91.76\\\cellcolor{LightGreen}+0.84} & \makecell{95.13\\\cellcolor{LightGreen}+0.14} & \makecell{91.92\\\cellcolor{LightGreen}+0.32} & \makecell{93.53\\\cellcolor{LightGreen}+0.38} \\
\midrule
ECL-GSR & 
\makecell{93.32\\\cellcolor{LightGreen}+0.62} & \makecell{84.75\\\cellcolor{LightRed}-4.16} & \makecell{97.91\\\cellcolor{LightGreen}+0.12} & \makecell{94.60\\\cellcolor{LightRed}-0.1} & \makecell{\textbf{92.59}\\\cellcolor{LightGreen}\textbf{+1.67}} & \makecell{\textbf{95.96}\\\cellcolor{LightGreen}\textbf{+0.97}} & \makecell{91.88\\\cellcolor{LightGreen}+0.28} & \makecell{92.94\\\cellcolor{LightRed}-0.2} \\
\midrule
\MissionHD\ & 
\makecell{\textbf{93.41}\\\cellcolor{LightGreen}\textbf{+0.71}} & \makecell{\textbf{91.07}\\\cellcolor{LightGreen}\textbf{+2.16}} & \makecell{98.32\\\cellcolor{LightGreen}+0.53} & \makecell{94.45\\\cellcolor{LightRed}-0.25} & \makecell{91.29\\\cellcolor{LightGreen}+0.37} & \makecell{95.16\\\cellcolor{LightGreen}+0.17} & \makecell{\textbf{91.95}\\\cellcolor{LightGreen}\textbf{+0.35}} & \makecell{\textbf{93.71}\\\cellcolor{LightGreen}\textbf{+0.56}} \\
\bottomrule
\end{tabular}
}
\end{table}

\begin{table*}[t]
\centering
\caption{
AUC (\%) for VAD (Normal) and VAR on the UCF-Crime dataset. Deltas (relative to the \texttt{defaultMSGs}) are shown below each main value, with green indicating improvement and red indicating degradation.
}
\label{tab:ucf_ab_numrefines_thresholds}
\resizebox{\textwidth}{!}{
\begin{tabular}{cc|c|ccccccccccccc|c}
\toprule
\textbf{\#Refine} & \textbf{Threshold} & \textbf{Normal} & \textbf{Abuse} & \textbf{Arrest} & \textbf{Arson} & \textbf{Assault} & \textbf{Burglary} & \textbf{Explosion} & \textbf{Fighting} & \textbf{RoadAcc.} & \textbf{Robbery} & \textbf{Shoplifting} & \textbf{Shooting} & \textbf{Stealing} & \textbf{Vandalism} & \textbf{mAUC} \\
\midrule
\multicolumn{2}{c|}{defaultMSGs} & 
92.14 & 65.59 & 81.47 & 89.45 & 27.42 & 52.30 & 72.67 & 90.48 & 95.11 & 47.64 & 92.88 & 67.58 & 88.60 & 76.58 & 72.91 \\
\cmidrule(lr){1-17}
\multirow{8}{*}{1} 
& 0.1 & 
\makecell{89.82\\\cellcolor{LightRed}-2.32} & \makecell{87.78\\\cellcolor{LightGreen}+22.19} & \makecell{84.08\\\cellcolor{LightGreen}+2.61} & \makecell{89.81\\\cellcolor{LightGreen}+0.36} & \makecell{57.88\\\cellcolor{LightGreen}+30.46} & \makecell{81.28\\\cellcolor{LightGreen}+28.98} & \makecell{83.77\\\cellcolor{LightGreen}+11.10} & \makecell{90.74\\\cellcolor{LightGreen}+0.26} & \makecell{96.49\\\cellcolor{LightGreen}+1.38} & \makecell{58.30\\\cellcolor{LightGreen}+10.66} & \makecell{95.16\\\cellcolor{LightGreen}+2.28} & \makecell{78.75\\\cellcolor{LightGreen}+11.17} & \makecell{89.42\\\cellcolor{LightGreen}+0.82} & \makecell{78.62\\\cellcolor{LightGreen}+2.04} & \makecell{82.47\\\cellcolor{LightGreen}+9.56} \\
\cmidrule(lr){2-17}
& 0.2 & 
\makecell{95.67\\\cellcolor{LightGreen}+3.53} & \makecell{88.30\\\cellcolor{LightGreen}+22.71} & \makecell{84.63\\\cellcolor{LightGreen}+3.16} & \makecell{89.19\\\cellcolor{LightRed}-0.26} & \makecell{47.09\\\cellcolor{LightGreen}+19.67} & \makecell{83.19\\\cellcolor{LightGreen}+30.89} & \makecell{78.29\\\cellcolor{LightGreen}+5.62} & \makecell{91.42\\\cellcolor{LightGreen}+0.94} & \makecell{97.58\\\cellcolor{LightGreen}+2.47} & \makecell{62.55\\\cellcolor{LightGreen}+14.91} & \makecell{93.24\\\cellcolor{LightGreen}+0.36} & \makecell{72.32\\\cellcolor{LightGreen}+4.74} & \makecell{92.23\\\cellcolor{LightGreen}+3.63} & \makecell{93.30\\\cellcolor{LightGreen}+16.72} & \makecell{82.56\\\cellcolor{LightGreen}+9.66} \\
\cmidrule(lr){2-17}
& 0.4 & 
\makecell{92.90\\\cellcolor{LightGreen}+0.76} & \makecell{74.92\\\cellcolor{LightGreen}+9.33} & \makecell{81.86\\\cellcolor{LightGreen}+0.39} & \makecell{85.91\\\cellcolor{LightRed}-3.54} & \makecell{30.80\\\cellcolor{LightGreen}+3.38} & \makecell{74.75\\\cellcolor{LightGreen}+22.45} & \makecell{71.38\\\cellcolor{LightRed}-1.29} & \makecell{90.91\\\cellcolor{LightGreen}+0.43} & \makecell{94.93\\\cellcolor{LightRed}-0.18} & \makecell{56.07\\\cellcolor{LightGreen}+8.43} & \makecell{92.51\\\cellcolor{LightRed}-0.37} & \makecell{53.42\\\cellcolor{LightRed}-14.16} & \makecell{91.66\\\cellcolor{LightGreen}+3.06} & \makecell{76.19\\\cellcolor{LightRed}-0.39} & \makecell{75.02\\\cellcolor{LightGreen}+2.12} \\
\cmidrule(lr){2-17}
& 0.8 & 
\makecell{92.12\\\cellcolor{LightRed}-0.02} & \makecell{76.69\\\cellcolor{LightGreen}+11.10} & \makecell{80.52\\\cellcolor{LightRed}-0.95} & \makecell{93.60\\\cellcolor{LightGreen}+4.15} & \makecell{44.43\\\cellcolor{LightGreen}+17.01} & \makecell{86.60\\\cellcolor{LightGreen}+34.30} & \makecell{77.11\\\cellcolor{LightGreen}+4.44} & \makecell{92.70\\\cellcolor{LightGreen}+2.22} & \makecell{96.69\\\cellcolor{LightGreen}+1.58} & \makecell{59.38\\\cellcolor{LightGreen}+11.74} & \makecell{95.43\\\cellcolor{LightGreen}+2.55} & \makecell{74.33\\\cellcolor{LightGreen}+6.75} & \makecell{90.88\\\cellcolor{LightGreen}+2.28} & \makecell{80.14\\\cellcolor{LightGreen}+3.56} & \makecell{80.65\\\cellcolor{LightGreen}+7.75} \\
\cmidrule(lr){1-17}
\multirow{8}{*}{2} 
& 0.1 & 
\makecell{94.16\\\cellcolor{LightGreen}+2.02} & \makecell{63.67\\\cellcolor{LightRed}-1.92} & \makecell{76.06\\\cellcolor{LightRed}-5.41} & \makecell{81.74\\\cellcolor{LightRed}-7.71} & \makecell{66.25\\\cellcolor{LightGreen}+38.83} & \makecell{79.15\\\cellcolor{LightGreen}+26.85} & \makecell{75.14\\\cellcolor{LightGreen}+2.47} & \makecell{94.47\\\cellcolor{LightGreen}+3.99} & \makecell{91.84\\\cellcolor{LightRed}-3.27} & \makecell{53.09\\\cellcolor{LightGreen}+5.45} & \makecell{91.68\\\cellcolor{LightRed}-1.20} & \makecell{63.84\\\cellcolor{LightRed}-3.74} & \makecell{85.06\\\cellcolor{LightRed}-3.54} & \makecell{86.18\\\cellcolor{LightGreen}+9.60} & \makecell{77.55\\\cellcolor{LightGreen}+4.64} \\
\cmidrule(lr){2-17}
& 0.2 & 
\makecell{91.59\\\cellcolor{LightRed}-0.55} & \makecell{83.47\\\cellcolor{LightGreen}+17.88} & \makecell{74.81\\\cellcolor{LightRed}-6.66} & \makecell{88.28\\\cellcolor{LightRed}-1.17} & \makecell{78.37\\\cellcolor{LightGreen}+50.95} & \makecell{82.55\\\cellcolor{LightGreen}+30.25} & \makecell{66.85\\\cellcolor{LightRed}-5.82} & \makecell{89.95\\\cellcolor{LightRed}-0.53} & \makecell{94.69\\\cellcolor{LightRed}-0.42} & \makecell{50.72\\\cellcolor{LightGreen}+3.08} & \makecell{94.02\\\cellcolor{LightGreen}+1.14} & \makecell{73.45\\\cellcolor{LightGreen}+5.87} & \makecell{91.45\\\cellcolor{LightGreen}+2.85} & \makecell{93.49\\\cellcolor{LightGreen}+16.91} & \makecell{81.70\\\cellcolor{LightGreen}+8.79} \\
\cmidrule(lr){2-17}
& 0.4 & 
\makecell{93.37\\\cellcolor{LightGreen}+1.23} & \makecell{74.98\\\cellcolor{LightGreen}+9.39} & \makecell{84.24\\\cellcolor{LightGreen}+2.77} & \makecell{93.52\\\cellcolor{LightGreen}+4.07} & \makecell{23.28\\\cellcolor{LightRed}-4.14} & \makecell{71.83\\\cellcolor{LightGreen}+19.53} & \makecell{63.87\\\cellcolor{LightRed}-8.80} & \makecell{94.91\\\cellcolor{LightGreen}+4.43} & \makecell{95.32\\\cellcolor{LightGreen}+0.21} & \makecell{48.01\\\cellcolor{LightGreen}+0.37} & \makecell{94.23\\\cellcolor{LightGreen}+1.35} & \makecell{78.24\\\cellcolor{LightGreen}+10.66} & \makecell{74.21\\\cellcolor{LightRed}-14.39} & \makecell{83.75\\\cellcolor{LightGreen}+7.17} & \makecell{75.41\\\cellcolor{LightGreen}+2.50} \\
\cmidrule(lr){2-17}
& 0.8 & 
\makecell{92.47\\\cellcolor{LightGreen}+0.33} & \makecell{83.86\\\cellcolor{LightGreen}+18.27} & \makecell{83.49\\\cellcolor{LightGreen}+2.02} & \makecell{93.10\\\cellcolor{LightGreen}+3.65} & \makecell{38.79\\\cellcolor{LightGreen}+11.37} & \makecell{77.67\\\cellcolor{LightGreen}+25.37} & \makecell{71.93\\\cellcolor{LightRed}-0.74} & \makecell{93.89\\\cellcolor{LightGreen}+3.41} & \makecell{95.64\\\cellcolor{LightGreen}+0.53} & \makecell{53.36\\\cellcolor{LightGreen}+5.72} & \makecell{94.74\\\cellcolor{LightGreen}+1.86} & \makecell{74.75\\\cellcolor{LightGreen}+7.17} & \makecell{90.32\\\cellcolor{LightGreen}+1.72} & \makecell{82.05\\\cellcolor{LightGreen}+5.47} & \makecell{79.51\\\cellcolor{LightGreen}+6.60} \\
\bottomrule
\end{tabular}
}
\end{table*}

\begin{table}[t]
\centering
\caption{
AUC (\%) for VAD (Normal) and VAR on the XD-Violence dataset, only using a single-time refined MSGs. Deltas (relative to the \texttt{defaultMSGs}) are shown below each main value, with green indicating improvement and red indicating degradation.
}
\label{tab:xdv_ab_numrefines_thresholds}
\resizebox{0.5\textwidth}{!}{
\begin{tabular}{c|c|cccccc|c}
\toprule
\textbf{Threshold} & \textbf{Normal} & \textbf{Abuse} & \textbf{CarAccident} & \textbf{Explosion} & \textbf{Fighting} & \textbf{Riot} & \textbf{Shooting} & \textbf{mAUC} \\
\midrule
\multicolumn{1}{c|}{defaultMSGs} & 
92.70 & 88.91 & 97.79 & 94.70 & 90.92 & 94.99 & 91.60 & 93.15 \\
\midrule
0.1 & 
\makecell{93.34\\\cellcolor{LightGreen}+0.64} & \makecell{90.84\\\cellcolor{LightGreen}+1.93} & \makecell{98.05\\\cellcolor{LightGreen}+0.26} & \makecell{94.52\\\cellcolor{LightRed}-0.18} & \makecell{91.52\\\cellcolor{LightGreen}+0.60} & \makecell{95.52\\\cellcolor{LightGreen}+0.53} & \makecell{91.73\\\cellcolor{LightGreen}+0.13} & \makecell{93.70\\\cellcolor{LightGreen}+0.55} \\
\midrule
0.2 & 
\makecell{93.41\\\cellcolor{LightGreen}+0.71} & \makecell{91.07\\\cellcolor{LightGreen}+2.16} & \makecell{98.32\\\cellcolor{LightGreen}+0.53} & \makecell{94.45\\\cellcolor{LightRed}-0.25} & \makecell{91.29\\\cellcolor{LightGreen}+0.37} & \makecell{95.16\\\cellcolor{LightGreen}+0.17} & \makecell{91.95\\\cellcolor{LightGreen}+0.35} & \makecell{93.71\\\cellcolor{LightGreen}+0.56} \\
\midrule
0.4 & 
\makecell{92.57\\\cellcolor{LightRed}-0.13} & \makecell{90.53\\\cellcolor{LightGreen}+1.62} & \makecell{97.88\\\cellcolor{LightRed}-0.09} & \makecell{94.78\\\cellcolor{LightGreen}+0.08} & \makecell{91.18\\\cellcolor{LightGreen}+0.26} & \makecell{95.79\\\cellcolor{LightGreen}+0.80} & \makecell{90.99\\\cellcolor{LightRed}-0.61} & \makecell{93.48\\\cellcolor{LightGreen}+0.37} \\
\midrule
0.8 & 
\makecell{93.09\\\cellcolor{LightGreen}+0.39} & \makecell{90.78\\\cellcolor{LightGreen}+1.87} & \makecell{97.91\\\cellcolor{LightGreen}+0.12} & \makecell{94.92\\\cellcolor{LightGreen}+0.22} & \makecell{91.10\\\cellcolor{LightGreen}+0.18} & \makecell{95.54\\\cellcolor{LightGreen}+0.55} & \makecell{91.59\\\cellcolor{LightRed}-0.01} & \makecell{93.64\\\cellcolor{LightGreen}+0.49} \\
\bottomrule
\end{tabular}
}
\end{table}

\subsection{Experimental Settings}
\label{sec:exp_settings}

We evaluate our proposed \MissionHD\ framework on the challenging tasks of VAD and VAR with an NVIDIA A100 GPU. Our experimental setup closely follows the methodology established in the previous paper~\cite{yun2025missiongnn}. We conduct experiments on two widely-used benchmarks for real-world anomaly detection: \textbf{UCF-Crime}~\cite{sultani2018real} and \textbf{XD-Violence}~\cite{xdv}, both of which consist of videos captured from surveillance cameras. For pre-trained joint embedding model, we used ImageBind-Huge~\cite{girdhar2023imagebind}. Also, GPT-4 is used to generate LLM-generated MSGs. Performance is measured using the Area Under the Receiver Operating Characteristic Curve (AUC) for individual classes and the mean AUC (mAUC) across all anomaly classes to provide a comprehensive assessment. Key hyperparameters for our model include a hyperdimensional space of $D=10,000$, a learnable dimensionality of $d=32$ for the learnable hypervectors, a hidden dimensionality of 32 for the final decision transformer layer, and a learning rate of $10^{-5}$. We explore graph refinement thresholds of 0.1, 0.2, 0.4, and 0.8. While our primary evaluation focuses on VAD/VAR, the \MissionHD\ framework is designed to be task-agnostic and can be readily applied to other domains requiring structured, interpretable reasoning over learned knowledge graphs.

\subsection{Quantitative Evaluation of \MissionHD}

We quantitatively evaluated our method against existing graph refinement approaches by applying them to LLM-generated MSGs. The baselines include four strong methods: direct optimization of the adjacency matrix, two graph neural networks (GCN~\cite{kipf2017semi} and GAT~\cite{velivckovic2017graph}), and the state-of-the-art energy-based refinement method ECL-GSR~\cite{zeng2025graph}. 

For adjacency matrix optimization, we used a learnable adjacency matrix and treated each value as an edge weight during downstream task training. For GCN and GAT, we encoded graphs for the downstream task and refined the adjacency matrix by measuring cosine similarity between learned node embeddings. For ECL-GSR, which has its own training loss, we treated each input video frame as a node connected to the MSGs and trained the model to learn graph distributions. Refined graphs were collected under varying thresholds, and we selected those corresponding to the threshold that yielded the highest mAUC improvement.

Results on UCF-Crime and XD-Violence are reported in \autoref{tab:vad_var_auc_compact} and \autoref{tab:xdv_auc_results}. ``None'' denotes the performance of LLM-generated graphs without refinement. All methods improve performance, confirming that raw LLM-generated MSGs are suboptimal and benefit from refinement. Our approach achieves the largest gains in both AUC and mAUC across both datasets, demonstrating its effectiveness compared to strong baselines.

\subsection{Performance under Different Thresholds and Number of Refinements}
\label{sec:performance_analysis}

We evaluated \MissionHD\ under varying thresholds and refinement rounds to assess its robustness. As shown in \autoref{tab:ucf_ab_numrefines_thresholds} and \autoref{tab:xdv_ab_numrefines_thresholds}, our refinement method consistently outperforms unrefined MSGs on both UCF-Crime and XD-Violence. A single refinement round yields substantial gains, including up to +9.66\% mAUC improvement on UCF-Crime. These consistent improvements across thresholds indicate that our task-driven HDC encoding–decoding process produces task-aligned graph structures that improve downstream reasoning.

We further tested multiple refinement rounds by feeding the output of one stage into the next. As shown in \autoref{tab:ucf_ab_numrefines_thresholds}, while some classes (e.g., \textit{Assault}) benefit, overall mAUC does not increase beyond the first refinement. We attribute this to the expressive capacity of the learnable graph edit hypervector: for the tested graph structures, a single hypervector suffices to capture the structural modifications needed to align with downstream data. This shows that one round of HDC-GSR is highly effective, and additional rounds provide limited returns, underscoring the efficiency of our method.

\subsection{Ablation Study on Dimensionalities}

We ablated two dimensionalities: hyperspace dimensionality $D$ and latent space dimensionality $d$ of the learnable hypervectors. \autoref{fig:ablation_dim} reports results on UCF-Crime with $D \in \{2500, 5000, 10000\}$ and $d \in \{16, 32, 64\}$. Performance improves with higher dimensionality. Increasing $D$ reduces decoding noise and improves edge score consistency, as predicted by our theoretical analysis, leading to better task-aligned MSGs. Increasing $d$ yields even larger gains than $D$, suggesting that higher latent dimensionality provides more expressive control over hyperdimensional representations and facilitates finding optimal hypervector encodings of the graph in HDC space.

\subsection{Empirical Study on Performance Gain Guarantee}

To test whether our method produces task-aligned refinements regardless of the initial MSGs, we applied random augmentations to LLM-generated MSGs. Given an initial edge density $m \in (0, 1)$, we randomly sampled $m|V^{(i)}||V^{(i+1)}|$ pairs from each neighboring node set $V^{(i)} \times V^{(i+1)}$ to form the augmented graph. We then compared downstream performance before and after applying our refinement method. 

Results on UCF-Crime are shown in \autoref{fig:edgedensity_mAUCs} and \autoref{fig:edgedensity_AUCs}, reporting VAR and class-wise VAD score improvements under different thresholds and edge densities ($0.25$, $0.3$, $0.5$, $0.7$, $0.75$). Two key findings emerge. First, improvements on randomly augmented MSGs are larger than those on LLM-generated MSGs, confirming that LLM outputs are suboptimal. Second, refinement performs best around threshold $0.5$. While our method consistently yields positive mAUC improvements across all thresholds, threshold $0.4$ provides the most stable gains across both classes and initial MSGs. These results demonstrate the robustness of our method against varying initial MSGs, showing that it reliably refines graphs toward more optimal task-aligned representations.

%% file: Sections/6_Conclusions.tex
We presented \MissionHD, an HDC-GSR framework for refining LLM-generated graphs in video anomaly detection and recognition. By optimizing task-aligned graph codes and decoding them into explicit structures, our method prunes irrelevant abstractions, strengthens task-relevant connections, and achieves consistent gains on UCF-Crime and XD-Violence. Our analysis also connects the framework to GNN path-sum formulations and establishes decoding guarantees, supporting the design of our encoding–decoding scheme. Overall, \MissionHD\ offers a principled and effective solution for adapting distribution-deficient reasoning graphs to downstream tasks.

\textbf{Limitations and Future Work.} Limitations include the current focus on layered DAGs and reliance on thresholding strategies. Future work will extend \MissionHD\ to more general graph types and automate refinement.

%% file: ref.bib
@String(CVPR= {IEEE Conf. Comput. Vis. Pattern Recog.})

@String(ECCV= {Eur. Conf. Comput. Vis.})

@String(ICLR = {Int. Conf. Learn. Represent.})

@String(AAAI = {AAAI})

@String(CVPR  = {CVPR})

@String(ECCV  = {ECCV})

@String(ICLR  = {ICLR})

@inproceedings{yun2025missiongnn,
  title={Missiongnn: Hierarchical multimodal gnn-based weakly supervised video anomaly recognition with mission-specific knowledge graph generation},
  author={Yun, Sanggeon and Masukawa, Ryozo and Na, Minhyoung and Imani, Mohsen},
  booktitle={2025 IEEE/CVF Winter Conference on Applications of Computer Vision (WACV)},
  pages={4736--4745},
  year={2025},
  organization={IEEE}
}

@inproceedings{rtfm,
  title={Weakly-supervised video anomaly detection with robust temporal feature magnitude learning},
  author={Tian, Yu and Pang, Guansong and Chen, Yuanhong and Singh, Rajvinder and Verjans, Johan W and Carneiro, Gustavo},
  booktitle={Proceedings of the IEEE/CVF international conference on computer vision},
  pages={4975--4986},
  year={2021}
}

@article{anomalyclip,
  title={Delving into clip latent space for video anomaly recognition},
  author={Zanella, Luca and Liberatori, Benedetta and Menapace, Willi and Poiesi, Fabio and Wang, Yiming and Ricci, Elisa},
  journal={Computer Vision and Image Understanding},
  volume={249},
  pages={104163},
  year={2024},
  publisher={Elsevier}
}

@InProceedings{vad1_cvpr,
    author    = {Yang, Zhiwei and Liu, Jing and Wu, Peng},
    title     = {Text Prompt with Normality Guidance for Weakly Supervised Video Anomaly Detection},
    booktitle = {Proceedings of the IEEE/CVF Conference on Computer Vision and Pattern Recognition (CVPR)},
    month     = {June},
    year      = {2024},
    pages     = {18899-18908}
}

@inproceedings{vad2_cvpr,
  title={Unbiased multiple instance learning for weakly supervised video anomaly detection},
  author={Lv, Hui and Yue, Zhongqi and Sun, Qianru and Luo, Bin and Cui, Zhen and Zhang, Hanwang},
  booktitle={Proceedings of the IEEE/CVF conference on computer vision and pattern recognition},
  pages={8022--8031},
  year={2023}
}

@inproceedings{masuakwa2025pv,
  title={PV-VTT: A Privacy-Centric Dataset for Mission-Specific Anomaly Detection and Natural Language Interpretation},
  author={Masuakwa, Ryozo and Yun, Sanggeon and Yamaguchi, Yoshiki and Imani, Mohsen},
  booktitle={2025 IEEE/CVF Winter Conference on Applications of Computer Vision (WACV)},
  pages={6415--6424},
  year={2025},
  organization={IEEE}
}

@article{poduval2022graphd,
  title={Graphd: Graph-based hyperdimensional memorization for brain-like cognitive learning},
  author={Poduval, Prathyush and Alimohamadi, Haleh and Zakeri, Ali and Imani, Farhad and Najafi, M Hassan and Givargis, Tony and Imani, Mohsen},
  journal={Frontiers in Neuroscience},
  volume={16},
  pages={757125},
  year={2022},
  publisher={Frontiers Media SA}
}

@inproceedings{luo2024rog,
title={Reasoning on Graphs: Faithful and Interpretable Large Language Model Reasoning},
author={Luo, Linhao and Li, Yuan-Fang and Haffari, Gholamreza and Pan, Shirui},
booktitle={International Conference on Learning Representations},
  year={2024}
}

@inproceedings{zhong2019graph_mil_video,
  title={Graph convolutional label noise cleaner: Train a plug-and-play action classifier for anomaly detection},
  author={Zhong, Jia-Xing and Li, Nannan and Kong, Weijie and Liu, Shan and Li, Thomas H and Li, Ge},
  booktitle={Proceedings of the IEEE/CVF conference on computer vision and pattern recognition},
  pages={1237--1246},
  year={2019}
}

@inproceedings{vqa-gnn,
  title={Vqa-gnn: Reasoning with multimodal knowledge via graph neural networks for visual question answering},
  author={Wang, Yanan and Yasunaga, Michihiro and Ren, Hongyu and Wada, Shinya and Leskovec, Jure},
  booktitle={Proceedings of the IEEE/CVF international conference on computer vision},
  pages={21582--21592},
  year={2023}
}

@inproceedings{huang2025building,
  title={Building a Mind Palace: Structuring Environment-Grounded Semantic Graphs for Effective Long Video Analysis with LLMs},
  author={Huang, Zeyi and Ji, Yuyang and Wang, Xiaofang and Mehta, Nikhil and Xiao, Tong and Lee, Donghyun and Vanvalkenburgh, Sigmund and Zha, Shengxin and Lai, Bolin and Yu, Licheng and others},
  booktitle={Proceedings of the Computer Vision and Pattern Recognition Conference},
  pages={24169--24179},
  year={2025}
}

@inproceedings{sharma2024vision,
  title={A vision check-up for language models},
  author={Sharma, Pratyusha and Shaham, Tamar Rott and Baradad, Manel and Fu, Stephanie and Rodriguez-Munoz, Adrian and Duggal, Shivam and Isola, Phillip and Torralba, Antonio},
  booktitle={Proceedings of the IEEE/CVF Conference on Computer Vision and Pattern Recognition},
  pages={14410--14419},
  year={2024}
}

@inproceedings{xdv,
    title={Not only Look, but also Listen: Learning Multimodal Violence Detection under Weak Supervision},
    author={Wu, Peng and Liu, jing and Shi, Yujia and Sun, Yujia and Shao, Fangtao and Wu, Zhaoyang and Yang, Zhiwei},
    booktitle={European Conference on Computer Vision (ECCV)},
    year={2020}
}

@inproceedings{ecl-gsr,
  title={Graph Structure Refinement with Energy-based Contrastive Learning},
  author={Zeng, Xianlin and Wang, Yufeng and Sun, Yuqi and Guo, Guodong and Ding, Wenrui and Zhang, Baochang},
  booktitle={Proceedings of the AAAI Conference on Artificial Intelligence},
  volume={39},
  number={21},
  pages={22326--22335},
  year={2025}
}

@inproceedings{gsr_23,
  title={Self-supervised graph structure refinement for graph neural networks},
  author={Zhao, Jianan and Wen, Qianlong and Ju, Mingxuan and Zhang, Chuxu and Ye, Yanfang},
  booktitle={Proceedings of the sixteenth ACM international conference on web search and data mining},
  pages={159--167},
  year={2023}
}

@article{kanerva2009hyperdimensional,
  title={Hyperdimensional computing: An introduction to computing in distributed representation with high-dimensional random vectors},
  author={Kanerva, Pentti},
  journal={Cognitive computation},
  volume={1},
  pages={139--159},
  year={2009},
  publisher={Springer}
}

@inproceedings{xu2017scenegraph,
  title={Scene Graph Generation by Iterative Message Passing},
  author={Xu, Danfei and Zhu, Yuke and Choy, Christopher and Fei-Fei, Li},
  booktitle={Computer Vision and Pattern Recognition (CVPR)},
  year={2017}
}

@inproceedings{jeong2025exploiting,
  title={Exploiting boosting in hyperdimensional computing for enhanced reliability in healthcare},
  author={Jeong, SungHeon and Barkam, Hamza Errahmouni and Yun, Sanggeon and Kim, Yeseong and Angizi, Shaahin and Imani, Mohsen},
  booktitle={2025 Design, Automation \& Test in Europe Conference (DATE)},
  pages={1--7},
  year={2025},
  organization={IEEE}
}

@article{thomas2021theoretical,
  title={A theoretical perspective on hyperdimensional computing},
  author={Thomas, Anthony and Dasgupta, Sanjoy and Rosing, Tajana},
  journal={Journal of Artificial Intelligence Research},
  volume={72},
  pages={215--249},
  year={2021}
}

@article{chen2024hdreason,
  title={Hdreason: Algorithm-hardware codesign for hyperdimensional knowledge graph reasoning},
  author={Chen, Hanning and Ni, Yang and Zakeri, Ali and Zou, Zhuowen and Yun, Sanggeon and Wen, Fei and Khaleghi, Behnam and Srinivasa, Narayan and Latapie, Hugo and Imani, Mohsen},
  journal={arXiv preprint arXiv:2403.05763},
  year={2024}
}

@inproceedings{li-etal-2025-graphotter,
    title = "{G}raph{OTTER}: Evolving {LLM}-based Graph Reasoning for Complex Table Question Answering",
    author = "Li, Qianlong  and
      Huang, Chen  and
      Li, Shuai  and
      Xiang, Yuanxin  and
      Xiong, Deng  and
      Lei, Wenqiang",
    editor = "Rambow, Owen  and
      Wanner, Leo  and
      Apidianaki, Marianna  and
      Al-Khalifa, Hend  and
      Eugenio, Barbara Di  and
      Schockaert, Steven",
    booktitle = "Proceedings of the 31st International Conference on Computational Linguistics",
    month = jan,
    year = "2025",
    address = "Abu Dhabi, UAE",
    publisher = "Association for Computational Linguistics",
    url = "https://aclanthology.org/2025.coling-main.368/",
    pages = "5486--5506"
}

@inproceedings{qiu2025step,
  title={STEP: Enhancing Video-LLMs' Compositional Reasoning by Spatio-Temporal Graph-guided Self-Training},
  author={Qiu, Haiyi and Gao, Minghe and Qian, Long and Pan, Kaihang and Yu, Qifan and Li, Juncheng and Wang, Wenjie and Tang, Siliang and Zhuang, Yueting and Chua, Tat-Seng},
  booktitle={Proceedings of the Computer Vision and Pattern Recognition Conference},
  pages={3284--3294},
  year={2025}
}

@inproceedings{dalvi2025hyperdimensional,
  title={Hyperdimensional representation learning for node classification and link prediction},
  author={Dalvi, Abhishek and Honavar, Vasant},
  booktitle={Proceedings of the Eighteenth ACM International Conference on Web Search and Data Mining},
  pages={88--97},
  year={2025}
}

@article{kleyko2022survey,
  title={A survey on hyperdimensional computing aka vector symbolic architectures, part i: Models and data transformations},
  author={Kleyko, Denis and Rachkovskij, Dmitri A and Osipov, Evgeny and Rahimi, Abbas},
  journal={ACM Computing Surveys},
  volume={55},
  number={6},
  pages={1--40},
  year={2022},
  publisher={ACM New York, NY}
}

@inproceedings{sultani2018real,
  title={Real-world anomaly detection in surveillance videos},
  author={Sultani, Waqas and Chen, Chen and Shah, Mubarak},
  booktitle={Proceedings of the IEEE conference on computer vision and pattern recognition},
  pages={6479--6488},
  year={2018}
}

@inproceedings{girdhar2023imagebind,
  title={Imagebind: One embedding space to bind them all},
  author={Girdhar, Rohit and El-Nouby, Alaaeldin and Liu, Zhuang and Singh, Mannat and Alwala, Kalyan Vasudev and Joulin, Armand and Misra, Ishan},
  booktitle={Proceedings of the IEEE/CVF conference on computer vision and pattern recognition},
  pages={15180--15190},
  year={2023}
}

@inproceedings{kipf2017semi,
  title={Semi-Supervised Classification with Graph Convolutional Networks},
  author={Kipf, Thomas N. and Welling, Max},
  booktitle={International Conference on Learning Representations (ICLR)},
  year={2017}
}

@article{velivckovic2017graph,
  title={Graph attention networks},
  author={Veli{\v{c}}kovi{\'c}, Petar and Cucurull, Guillem and Casanova, Arantxa and Romero, Adriana and Lio, Pietro and Bengio, Yoshua},
  journal={arXiv preprint arXiv:1710.10903},
  year={2017}
}

@inproceedings{zeng2025graph,
  title={Graph Structure Refinement with Energy-based Contrastive Learning},
  author={Zeng, Xianlin and Wang, Yufeng and Sun, Yuqi and Guo, Guodong and Ding, Wenrui and Zhang, Baochang},
  booktitle={Proceedings of the AAAI Conference on Artificial Intelligence},
  volume={39},
  number={21},
  pages={22326--22335},
  year={2025}
}
